\documentclass[11pt,draftcls,onecolumn,journal]{IEEEtran}
\usepackage{amsmath,amsfonts, amssymb}
\usepackage{algorithmic}
\usepackage{algorithm}
\usepackage{array}
\usepackage[caption=false,font=normalsize,labelfont=sf,textfont=sf]{subfig}
\usepackage{textcomp}
\usepackage{stfloats}
\usepackage{url}
\usepackage{verbatim}
\usepackage{bbm}
\usepackage{xcolor}

\usepackage{graphicx}
\usepackage{slashbox}
\usepackage{cite}
\usepackage{hyperref}
\usepackage{booktabs}
\usepackage{multirow}
\usepackage[framemethod=tikz]{mdframed}
\usepackage{threeparttable}
\usepackage{booktabs} 

\newtheorem{remark}{Remark}
\newtheorem{definition}{Definition}

\def \bX {{\mathbf{X}}}
\def \bx {{\mathbf{x}}}
\def \bs {{\mathbf{s}}}
\def \bz {{\mathbf{z}}}
\def \bZ {{\mathbf{Z}}}
\def \bh {{\mathbf{h}}}

\def \bC {{\mathbf{C}}}
\def \bQ {{\mathbf{Q}}}
\def \bS {{\mathbf{S}}}
\def \bV {{\mathbf{V}}}
\def \bK {{\mathbf{K}}}
\def \bA {{\mathbf{A}}}
\def \bO {{\mathbf{O}}}
\def \bM {{\mathbf{M}}}

\def \bV {{\mathbf{V}}}
\def \bU {{\mathbf{U}}}
\def \bs {{\mathbf{s}}}

\def \bB {{\mathbf{B}}}
\def \bW {{\mathbf{W}}}

\def \bH {{\mathbf{H}}}

\def \bu {{\mathbf{u}}}

\def \bI {{\mathbf{I}}}
\def \mR {{\mathbb{R}}}
\def \mE {{\mathbb{E}}}

\def \mZ {{\mathbb{Z}}}

\def \cH {{\mathcal{H}}}

\newtheorem{theorem}{Theorem}
\newcommand\blfootnote[1]{%
  \begingroup
  \renewcommand\thefootnote{}\footnote{#1}%
  \addtocounter{footnote}{-1}%
  \endgroup
}

\newcommand \blue[1]        {{\color{black}#1}}

\begin{document}

\title{Learning with Covariance Matrices \\ {\LARGE Principal Component Analysis Meets Learning with Graphs}}

\author{Saurabh Sihag, Andrea Cavallo, Elvin Isufi, Gonzalo Mateos, and Alejandro Ribeiro$^{\dagger}$

\thanks{$^{\dagger}$Saurabh Sihag is with the Department of Electrical and Computer Engineering at the University at Albany, SUNY, Albany, NY (email: \href{ssihag@albany.edu}{ssihag@albany.edu}). Andrea Cavallo (email: \href{A.Cavallo@tudelft.nl}{A.Cavallo@tudelft.nl}) and Elvin Isufi (email: \href{e.isufi-1@tudelft.nl}{e.isufi-1@tudelft.nl}) are with the Delft University of Technology, Delft, The Netherlands. Gonzalo Mateos is with the Department of Electrical and Computer Engineering at the
University of Rochester, Rochester, NY (email:\href{gmateosb@ece.rochester.edu}{gmateosb@ece.rochester.edu}). Alejandro Ribeiro is with the Department of Electrical and Systems Engineering at the University of Pennsylvania, Philadelphia, PA (email:\href{aribeiro@seas.upenn.edu}{aribeiro@seas.upenn.edu}).}
}

\markboth{IEEE Signal Processing Magazine}{IEEE Signal Processing Magazine}%


\maketitle

\blfootnote{This work was supported in part by the TU Delft AI Labs programme, NWO OTP GraSPA proposal $\#$19497, NWO VENI proposal 222.032, and by the SURE-AI Centre grant $\#$357482, Research Council of Norway.}

\section*{Introduction}
\label{sec:intro}

\begin{figure}[t]
    \centering
    \includegraphics[width=\linewidth]{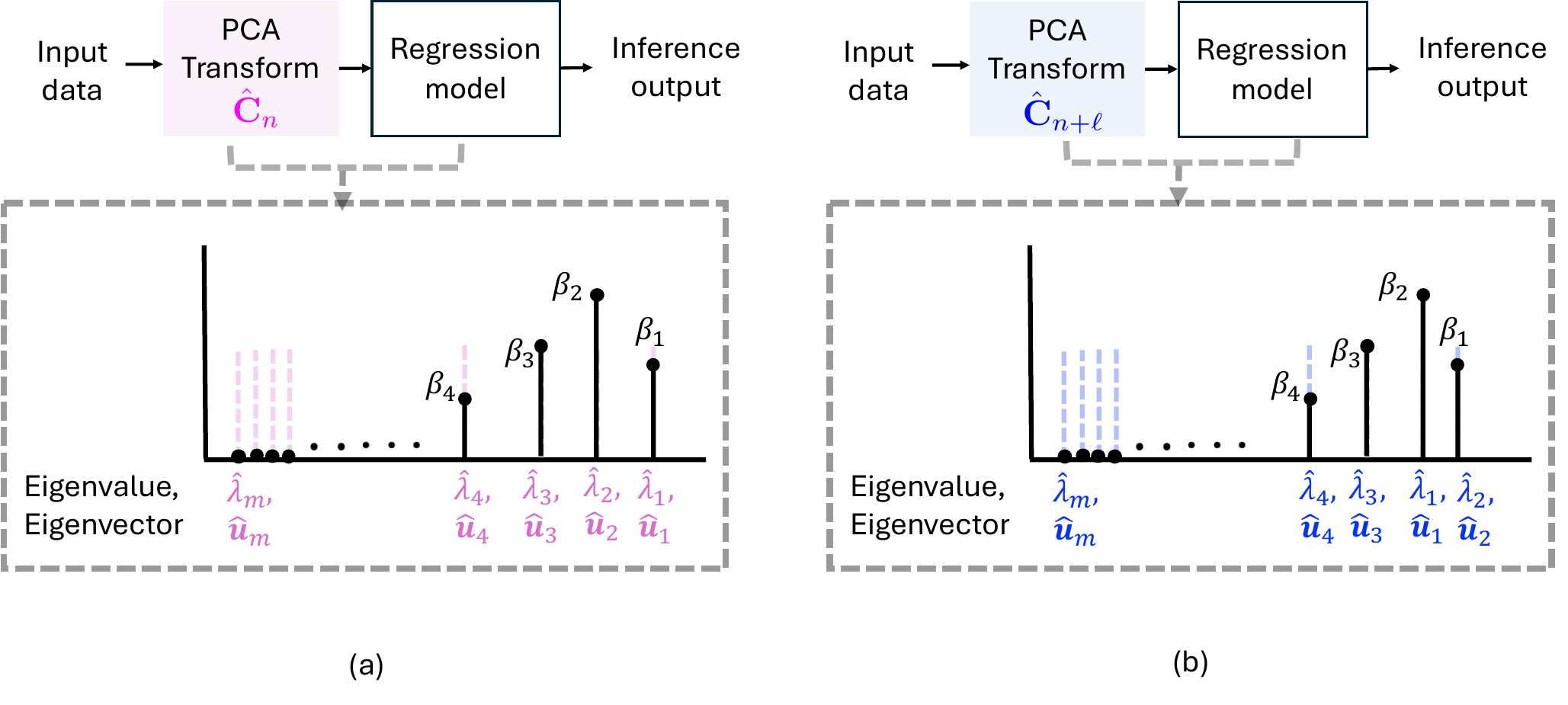}
    \caption{Potential instability of PCA-driven regression model illustrated through an example. (a) PCA-driven regression model consists of implementing a PCA transform based on $\hat\bC_n$ first, followed by a linear regression model. The learned coefficients $\{\beta_1,\dots,\beta_m\}$ of the linear model are determined from a given training dataset and represent the data-driven importance of individual eigenvectors of $\hat\bC_n$ to the learning objective. (b) Consider the scenario where the sample covariance matrix is re-estimated with $n+t$ samples to yield $\hat\bC_{n+t}$ and the learned coefficients $\{\beta_1,\dots,\beta_m\}$ are inherited from (a). If the estimate $\hat\lambda_1$ of first eigenvalue of $\bC$ in $\hat\bC_{n+t}$ is smaller than that of its estimate $\hat\lambda_2$ of the second eigenvalue of $\bC$ due to the finite sample effect, the coefficients $\beta_1$ and $\beta_2$ are associated with the estimates of eigenvectors $\bu_2$ and $\bu_1$, respectively. Consequently, the higher importance (in terms of $\beta_1$) placed on $\bu_2$ relative to $\bu_1$ in setting (b) is in contrast to (a). This may lead to potentially inconsistent regression performance relative to (a) as the eigenvectors of a covariance matrix are orthogonal to each other.}
    \label{fig:spm_stab}
\end{figure}

Covariance matrices encode linear (statistical) dependencies between different pairs of features within a dataset and have been the cornerstone of multivariate data analysis in a host of applications characterized by spatially distributed data acquisition protocols. Noteworthy examples with signal processing (SP) and machine learning (ML) relevance include neuroimaging~\cite{evans2013networks}, computer vision~\cite{murase1994illumination}, weather modeling~\cite{stephenson1997correlation}, traffic flow analysis~\cite{shao2014estimation}, and cloud computing~\cite{ismail2013detecting}, to name a few. Principal Component Analysis (PCA) is among the most widely adopted covariance matrix-driven statistical techniques~\cite{pcareview}. Specifically, PCA is an orthogonal linear transformation derived from the eigenvectors of the covariance matrix~\cite{shlens2014tutorial}. Because the eigenvectors and their corresponding eigenvalues of the covariance matrix are individually tied to the variance explained in a given dataset, PCA is often used as a dimensionality reduction technique that enjoys well-documented optimality properties. In a nutshell, one projects a dataset onto a lower-dimensional covariance eigenspace with the 
goal of preserving the 
``most signal'' 
for a given learning task. A schematic illustration of a prototypical PCA-driven regression model is provided in Fig.~\ref{fig:spm_stab} and Fig.~\ref{fig:pca_vf} (left) and consists of two major modules: (a) PCA transform on the input data that can be viewed as a (dimensionality-reducing) feature extractor; followed by (b) a linear regression model that assigns ``importance'' weights to individual principal components.  

In practice, all one has to work with are empirical covariance matrices estimated from data (referred to as sample covariance matrices), which are statistical estimates of the population covariance matrix. In this context, a first key challenge facing PCA-based learning approaches is the lack of reproducibility of findings~\cite{joliffe1992principal, jolliffe2016principal}, which stem from finite sample-induced stochastic perturbations in the eigenvector estimates associated with \textit{close} eigenvalues of the sample covariance matrix~\cite{loukas2017close}
; see also `Challenges to Learning with PCA' and Fig.~\ref{fig:spm_stab}. Moreover, conventional PCA-driven models are not adept at multiscale information processing needed for applications where data are acquired at different e.g., spatial scales; such as in geosciences, atmospheric sciences, power grids, and brain imaging~\cite{willsky2002multiresolution}. This is because as the number of input features changes, so does the size of the covariance matrix and the properties of its eigenspace -- which needs to be recomputed from scratch, leading also to computational issues.  
Hence, fundamental theoretical limits of inference across multiscale datasets (while effectively leveraging their redundancies) are categorically lacking. Furthermore, 
the PCA-based feature preprocessing step is decoupled from the downstream learning task. In `GSP-Driven Implementation of PCA', we pivot towards a graph-driven perspective to covariance matrix, which ties PCA to the domain of graph signal processing~\cite{leus2023graph}. This novel perspective provides an alternative, linear-shift-and-sum implementation in terms of the covariance matrix for PCA-driven learning models, while addressing the aforementioned challenges to its practical implementation and laying the groundwork for significant conceptual advancements to PCA. 

Leveraging covariance matrices as graph representations (for example, covariance matrices derived from functional brain measurements in network neuroscience; see Fig.~\ref{fig:cov_ex}) is ubiquitous in graph-driven machine learning frameworks~\cite{mateos2018spmag}. For instance, a graph neural network (GNN) can operationally exploit the graph structure manifested in the form of covariance matrices for learning. Recent advances in graph signal processing (GSP) have provided key operational and theoretical principles for GNNs~\cite{ruiz2021graph, gama2020graphs, gama2020stability, li2021graph}. We begin the theoretical discussions by elucidating the natural connection between PCA and theoretical understanding of the fundamental computational block within GSP: a graph convolutional filter that has a linear-shift-and-sum implementation in the matrix representation of a graph. Spectral analysis of the graph filter reveals that its outcome is tied to data-driven exploitation of the eigenspectrum of the graph~\cite{isufi2024graph}. Setting the graph to be a covariance matrix within a graph filter reveals an implementation of a learning model that is conceptually identical to PCA (Fig.~\ref{fig:pca_vf}). Building upon the said \underline{equivalence} between PCA-based learning and graph filters over covariance matrices, recent works have undertaken the mathematical study of GNNs with covariance matrices as graphs, termed \emph{coVariance neural networks (VNNs)}~\cite{sihag2022covariance, cavallo2024sparsecovarianceneuralnetworks, cavallo2024spatiotemporal, cavallo2024fair}. A key aim of this tutorial is to convey the broad scope of improvements and generalization capabilities offered by VNN models over PCA-based and other related learning pipelines. In this context, our aim is not to diminish the PCA transform or put VNNs in contrast to it, but rather to offer significant conceptual advancements to PCA through VNN models. 


In `Stability of VNNs', we discuss the theoretical results underlying the stability of VNN outcomes to stochastic perturbations in the sample covariance matrix. As an upshot of the theoretical analyses on the stability of VNNs, we also elucidate conditions under which VNNs offer guaranteed reproducibility of learning outcomes on independently collected datasets, leading to an inherently stable deep learning alternative to PCA-based pipelines. While existing studies on GNNs with graph convolutional filters have established stability to abstract (absolute or relative) perturbations~\cite{gama2020stability}, the \emph{stability results for VNNs} discussed herein are markedly refined by borrowing precise error models and mathematical tools from perturbation theory of covariance matrices~\cite{loukas2017close}.

In `Transferability of VNNs', we leverage the mathematical lens of graphons and limit objects of graph sequences to demonstrate that VNNs exhibit transference across multiscale datasets (datasets capturing the same information with distinct numbers of features or scales). Rigorous analysis of the \underline{transferability of VNNs across multiscale datasets} is key to understanding how the redundancy of information within covariance matrices of different dimensionalities could be fruitfully exploited for principled design and efficient deployment of deep NN models {in communications, array processing, and neuroimaging analysis systems.}

Recent advances in graph signal processing (GSP) and spectral theory of GNNs have provided key operational and theoretical principles for stability or transferability of GNNs~\cite{ruiz2021graph, gama2020graphs, gama2020stability, li2021graph}. However, there are critical theoretical gaps between the nuances of spatial and spectral information inherent to a covariance matrix and the known mathematical and empirical characteristics of GNNs~\cite{gama2020graphs}, thus, inhibiting \textit{principled} design and application of GNNs across broad domains where covariance matrices emerge as the go-to descriptor of multivariate data structure. \blue{This article lays the groundwork for rigorous theoretical studies of GNNs with covariance matrices as graphs, which have the potential to elucidate a principled understanding underlying the deployment of deep neural networks (NNs) across a wide spectrum of timely SP applications. Moreover, this article elucidates significant conceptual updates in terms of implementation and reproducibility to traditional PCA-driven inference models from the lens of GSP.} All in all, the surveyed theoretical analyses of VNN stability and transferability properties are rooted at the crossroads of statistical SP and (geometric) deep learning, offering an ideal framework to bridge the gap between traditional statistical approaches using covariance matrices and the recently developed insights and capabilities offered by GNNs.

\section*{Learning with Covariance Matrices: A PCA Perspective}

Covariance matrices encode pairwise, linear statistical dependencies in a given dataset. Formally, for a dataset consisting of $n$ random, independent, and identically distributed (i.i.d) data samples~$\bx_i \in \mR^{m },\forall i\in\{1,\dots,n\}$. Using the samples $\bx_i, \forall i\in\{1,\dots,n\}$, the empirical covariance matrix is 
\begin{align}\label{sample_cov}
    \hat \bC_{n} \triangleq \frac{1}{n-1} \sum\limits_{i=1}^n\bar\bx_i \bar\bx_i^{\sf T}\;,
\end{align}
\noindent
\blue{where $\bar\bx_i = \bx_i - \bar \bx$ and $\bar\bx$ is the sample mean.} The sample covariance matrix is a statistical estimate of the true covariance matrix (also referred to as ensemble covariance matrix) ${\bf C}$. This ensemble covariance matrix $\bC$ is determined from an $m-$dimensional random vector $\bx \in \mR^{m}$ as $\bC \triangleq \mE[(\bx-\mE[\bx])(\bx-\mE[\bx])^{\sf T}]$.  However, the data model for $\bx$ or its statistics are often not available in practical applications. The convergence between $\hat \bC_{n}$ and ${\bf C}$ is stochastic, with its principles governed by perturbation theory~\cite{loukas2017close}. For conciseness, we reuse the notation $\bx$ to denote a data sample from the given dataset in subsequent sections of this article. 

\subsection*{Principal Component Analysis}
PCA leverages the covariance matrix to reveal the hidden, simplified structure of complex data~\cite{shlens2014tutorial}. PCA has been used abundantly across many domains where multivariate datasets appear~\cite{jolliffe2016principal, pcareview}. The workhorse PCA is deployed as a linear transformation over the data samples, where the transformation is characterized by the eigenspectrum of the covariance matrix. Given the eigendecomposition 
\begin{align}\label{eigen}
    \hat{\bf C}_n = \hat{\bf U}\hat{\bf \Lambda} \hat{\bf U}^{\sf T}\;.
\end{align}
where $\hat{\bf U} = [\hat{\bf u}_1,\dots, \hat{\bf u}_m]$ is the orthonormal matrix of $m$ eigenvectors, and $\hat{\bf \Lambda} = {\sf diag}(\hat\lambda_1,\dots, \hat\lambda_m)$ is the diagonal matrix of eigenvalues ordered as $\hat\lambda_1\geq \hat\lambda_2\dots\geq \hat\lambda_m$, the PCA transformation of a data sample $\bx$ is given by
\begin{align}\label{pca1}
    \tilde \bx = \hat{\bf U}^{\sf T} {\bf x}\;.
\end{align}
\noindent
Thus, PCA is an orthogonal, linear transformation that leads to a change of basis of ${\bf x}$, with the new basis determined by the eigenvectors of $\hat\bC_n$. The eigenvectors of $\hat\bC_n$ are typically referred to as the \emph{principal components}. Because $\hat\bC_n$ is the estimate of $\bC$, its eigenvectors $\hat\bU$ and eigenvalues $\hat{\bf \Lambda}$ are estimates of the eigenvectors $\bU = [\bu_1,\dots, \bu_m]$ and eigenvalues ${\bf \Lambda} = {\sf diag}(\lambda_1,\dots, \lambda_m)$ \blue{of $\bC$}, respectively. In practice, we observe $\hat\bC_n$ and not $\bC$ and therefore, the PCA transform is defined in terms of $\hat\bU$. Furthermore, the eigenvalues of $\hat\bC_n$ encode the \emph{variance} of the dataset in the specific direction of the corresponding principal component. Specifically, it can be readily shown that~\cite{shlens2014tutorial}
\begin{align}
   \blue{\frac{1}{n-1} \sum\limits_{i=1}^n | \hat\bu_j^{\sf T}{\bf x}_i|^2 = \hat\lambda_j\;.}
 \end{align}
Thus, the variance of a dataset along the direction of a specific principal component is commensurate with the magnitude of its corresponding eigenvalue. Equivalently, it could be stated that the principal components \emph{explain} the variance within a given dataset in proportion to the magnitude of their corresponding eigenvalues.  

 PCA helps elucidate the hidden structure or patterns of interest in the raw data, which are then leveraged for further statistical analyses (such as clustering, regression, etc.). For instance, the framework for a PCA-driven regression model is illustrated in  Fig.\ref{fig:pca_vf}. PCA-driven approaches form the workhorse techniques for statistical analysis across a broad range of learning tasks and data modalities. 

\subsection*{Challenges to Learning with PCA}

Despite its widespread use, PCA is constrained by several significant limitations regarding reproducibility of inferred outcomes and computational bottlenecks in implementation. The challenges concerning the reproducibility of inferred outcomes arise from the fact that estimating principal components from finite samples is highly sensitive to noise and statistical uncertainty, particularly in low-data regimes or when covariance eigenvalues are closely spaced~\cite{joliffe1992principal}. According to the Davis-Kahan theorem~\cite{davis1970rotation}, the estimation error for the first $k$ principal components for a sample covariance matrix is inversely proportional to the eigengap $\lambda_k - \lambda_{k+1}$. Formally, for any two eigenvectors $\hat\bu_j$ and $\bu_i$ of $\hat\bC_n$ and $\bC$, respectively, their inner product scales with the number of samples $n$ and the eigengap $|\lambda_i - \lambda_j|$ as~\cite[Theorem 4.1]{loukas2017close}
\begin{align}\label{eiggap}
    \mathbb{P}\left(|\langle \hat\bu_i, \bu_j\rangle | \geq t\right) = {\cal O} \left(\frac{1}{n t^2(\lambda_i-\lambda_j)^2} \right) \;,
\end{align}
for some constant $t>0$. The relationship in~\eqref{eiggap} implies that as eigenvalues become less distinct, the corresponding eigenspaces are harder to separate, leading to inaccurate component estimation. Prior studies have proposed various approaches to tackle the instability arising from ill-defined principal components. The studies in~\cite{joliffe1992principal} and~\cite{szwagier2023curse} proposed a heuristic principled subspace-based approach to PCA based on the hypothesis that a subspace defined by a group of principal components may retain better stability relative to individual principal components if the eigenvalues corresponding to the said subspace are well-separated from the adjacent ones. PCA has also been studied with structural constraints on the covariance matrix (such as  sparsity~\cite{jankova2021biased} and diagonal structure~\cite{ke2020diagonally}), where the said constraints enable improved robustness of learning outcomes to close eigenvalues.

Furthermore, implementation of PCA requires explicit computation of the eigendecomposition of the sample covariance matrix, an $\mathcal{O}(m^3)$ operation that becomes computationally prohibitive for high-dimensional datasets. In this context, existing studies have proposed various computationally efficient approaches to implementing PCA, including identification of principal components under sparsity constraints~\cite{xie2024least} or rank of the covariance matrix~\cite{del2023sparse}.

Finally, the conventional PCA implementations lack both induction and transferability. For instance, changes in the dimensionality of the dataset necessitate a complete recomputation of the principal components, limiting its utility in dynamic or large-scale applications (for instance, datasets with multiple spatial scales; see Fig.~\ref{mscale}). In contrast to the prior studies that provide specialized solutions proposed to specific challenges faced by PCA, this article provides a holistic conceptual update to PCA through the lens of graph treatment of covariance matrix and GSP-driven information processing.

\begin{figure}
    \centering
    \includegraphics[width=\linewidth]{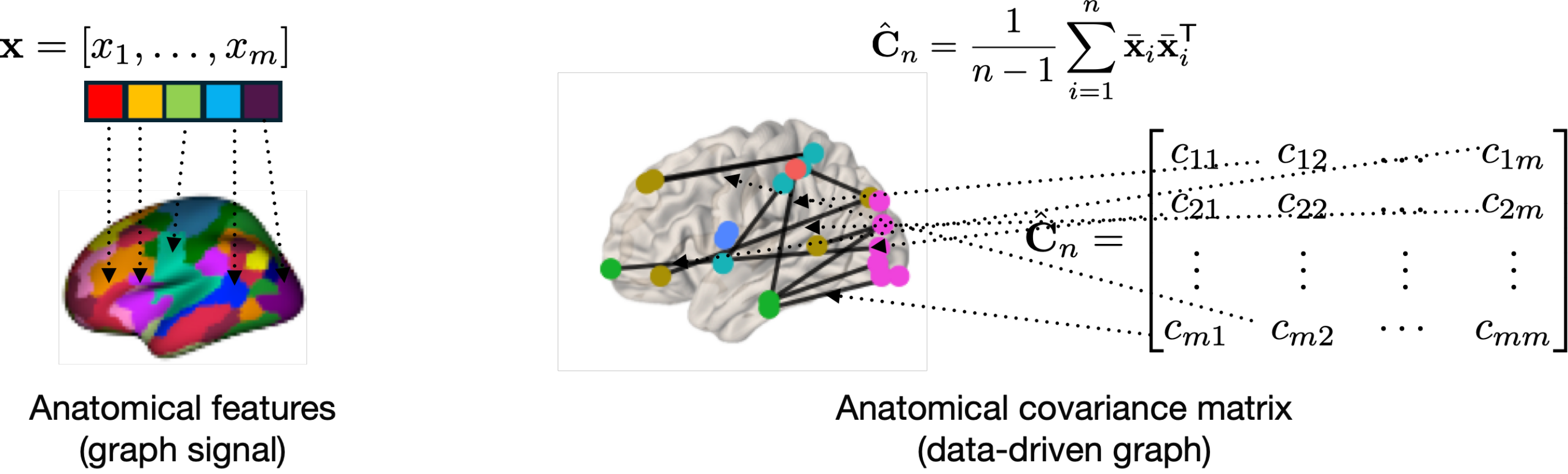}
    \caption{A graph perspective to covariance matrix in the context of neuroimaging data. The data vector $\bx$ represents anatomical information across the brain as it consists of readings or measurements at distinct brain regions. The non-diagonal entries of the corresponding (anatomical) covariance matrix represent the extent of linear coherence between the measurements at a given pair of brain regions. Thus, the covariance matrix admits a graph interpretation over the brain surface, with the nodes being different brain regions and the edges representing the coherence (for example, functional or anatomical) between them. }
    \label{fig:cov_ex}
\end{figure}

\section*{GSP-driven Implementation of PCA}
Principles of GSP have bridged signal processing insights and mathematical theory over graphs with the empirical successes of GNNs~\cite{ruiz2021graph}. A unique operational and conceptual perspective to PCA is obtained by treating covariance matrix as a graph. 

\noindent
{\bf Covariance matrix as a graph.} Formally, \emph{a graph} ${\cal G}({\cal V}, {\cal E}, {\cal W})$ is a triplet with a set of $m$ nodes ${\cal V} = \{1,\dots, m\} $, a set of undirected edges ${\cal E} \subseteq {\cal V}\times {\cal V} $, and an weight function ${\cal W}: {\cal E}\mapsto \mathbb{R}$. Furthermore, the graph topology can be compactly represented using a symmetric matrix of size $m\times m$, which encodes the edge and weight information. Within this framework, the constituent features of the $m$-dimensional data vector $\bx$ can be considered as `nodes' of a graph, whose graphical structure is represented by the covariance matrix $\hat\bC_n$ and $\bx$ being the signal over the graph (graph signal)~\cite{mateos2018spmag}. Figure~\ref{fig:cov_ex} provides a practical example where the graphical perspective of covariance matrix is meaningful in a network neuroscience context.

\begin{figure}
    \centering
    \includegraphics[width=\linewidth]{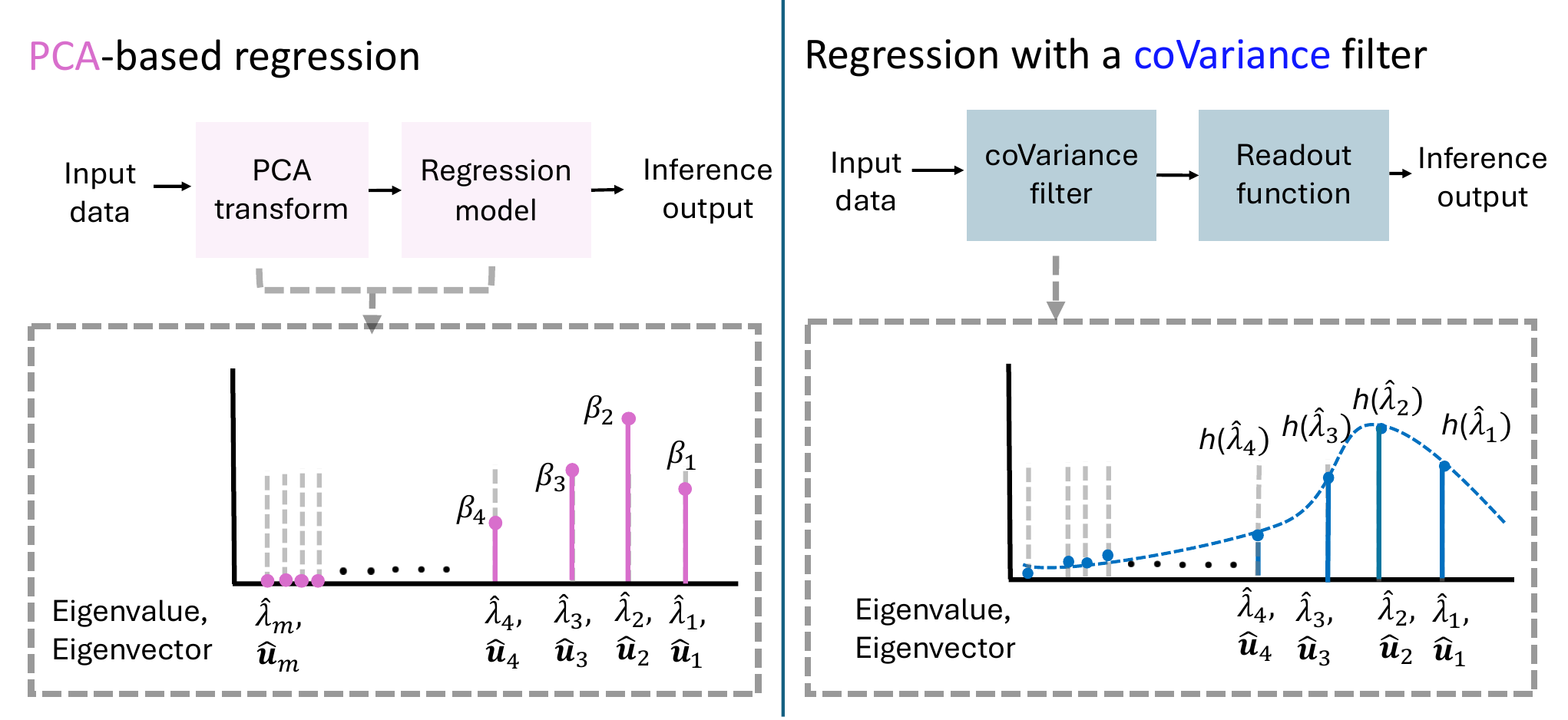}
    \caption{PCA-driven regression model versus a coVariance filter-based regression model.}
    \label{fig:pca_vf}
\end{figure}

\noindent
{\bf Graph Fourier Transform and PCA.} The graph Fourier transform (GFT) is the fundamental tool for spectral analysis of graph signals over a given graph structure. GFT of $\bx$ over $\hat\bC_n$ leverages the eigendecomposition in~\eqref{eigen} and is given by
\begin{align}\label{gft1}
    \breve \bx = \hat\bU^{\sf T}\bx\;.
\end{align}
Clearly, the GFT in~\eqref{gft1} reconciles with the PCA transform in~\eqref{pca1}. In the context of GSP, the eigenvalues of $\hat\bC_n$ are mathematical equivalent of the notion of graph frequencies in GSP~\cite{li2021graph}. Thus, the $i$-th entry of $ \breve \bx$, i.e., $[\breve \bx]_i$, represents the $i$-th Fourier coefficient corresponding to eigenvalue $\hat\lambda_i$. In an ML context, PCA-driven learning models learn the relative importance of individual eigenvectors of $\hat\bC_n$. For instance, the scalar output of the PCA-based regression model with a scalar response in Fig.~\ref{fig:pca_vf} is characterized by
\begin{align}
    \hat y &= \sum\limits_{i=1}^m \beta_i[\breve \bx]_i + \beta_0\;,\\
    &= \mathbf{1}^{\top} \bB\breve \bx + \beta_0\;,\label{pcaimp}
\end{align}
\blue{where $\mathbf{1}$ is an $m$-dimensional vector of all $1$s}, \blue{and $\bB = {\sf diag}(\beta_1,\dots,\beta_m)$} is the diagonal matrix of learned coefficients and $\beta_0\in \mathbb{R}$ is the bias term learned during training. In the GSP context, the operation $\bB\breve \bx$ in the implementation of PCA-driven regression model in~\eqref{pcaimp} is equivalent to implementing an `ideal' spectral filter, where the filter coefficients are given by the diagonal elements in~$\bB$. 

\noindent
{\bf Achieving Spectral Filtering with Polynomial  Operations.} Spectral filtering in the spirit of $\bB\breve \bx$ can be implemented via polynomial operations over $\hat\bC_n$. Consider the operation 
%
%
\begin{align}\label{gft2}
    \bz = (h_0 \bI + h_1\hat\bC_n)\bx\;,
\end{align}
where $h_0\in \mathbb{R}$ and $h_1\in \mathbb{R}$ are scalars. By leveraging the eigendecomposition of $\hat\bC_n$, we can rewrite $\bz$ in~\eqref{gft2} as
\begin{align}
    \bz &= (h_0 \bI + h_1\hat\bU {\bf \hat\Lambda} \hat\bU^\top) \bx  \\
    &= \hat\bU (h_0\hat{\bf\Lambda}^0) \hat\bU^\top \bx + \hat\bU (h_1\hat{\bf \Lambda}) \hat\bU^\top \bx\\
    &= \hat\bU (h_0\hat{\bf\Lambda}^0 + h_1\hat{\bf \Lambda}) \hat\bU^\top \bx\;.
\end{align}
Thus, the spectral analysis of the elementary operation $(h_0\bI + h_1\hat\bC_n)\bx$ \blue{in~\eqref{gft2} reveals that the dependence of output $\bz$ on eigenvector $\hat\bu_i$ is given by the function $h(\hat\lambda_i) = h_0 + h_1\hat\lambda_i$.} An increase in the sophistication of $h(\hat\lambda_i)$ (or equivalently, the design of the spectral filter) can readily be achieved by defining a higher-order polynomial in $\hat\bC_n$ and without the explicit eigendecomposition of $\hat\bC_n$. Specifically, for some arbitrary $K\in \mathbb{Z}^{+}$, the spectral analysis of the operation
\begin{align}\label{gft3}
    \bz = \sum\limits_{k=0}^K h_k \hat\bC_n^k\bx
\end{align}
yields 
\begin{align}\label{gft4}
    \bz &= \sum\limits_{k=0}^{K}h_k \hat\bU \hat{\bf \Lambda}^k \hat\bU^{\sf T} \bx \quad = \hat\bU \sum\limits_{k=0}^{K} h_k  \hat{\bf \Lambda}^k \hat\bU^{\sf T} \bx\;.
\end{align}
Thus, the dependence of the output $\bz$ in~\eqref{gft3} on eigenvector $\hat\bu_i$ is dictated by the filtering operation $h(\hat\lambda_i) = \sum\limits_{k=0}^Kh_k\hat\lambda_i^k$. The coefficients $\bh = [h_0, h_1,\dots, h_K]$ form the~\emph{learnable} parameters that are estimated from data. Thus, the relative importances of the principal components of $\hat\bC_n$ to the output $\bz$ in~\eqref{gft3} \blue{are} determined in a data-driven fashion (i.e., through functions $\{h({\hat\lambda_i})\}_{i=1}^m$). More specifically, the graph filter modifies the contribution of the $i$-th eigenvector of $\hat\bC_n$ via the function $h: \mathbb{R}\mapsto \mathbb{R}$ evaluated on the eigenvalue $\lambda_i$. Furthermore, in a spirit similar to that of PCA-based regression model in~\eqref{pcaimp}, a learning model that achieves learning objectives by data-driven exploitation of principal components can be realized via
\begin{align}
    \hat y = \mathbf{1}^{\top} \left(\sum\limits_{k=0}^{K} h_k \hat\bC_n^k\bx\right) \;.
\end{align}
The polynomial operation $\sum\limits_{k=0}^K h_k \hat\bC_n^k$ in~\eqref{gft3} is formally referred to as the `graph filter' operation in the GSP context~\cite{isufi2024graph} and denoted by $\bH(\hat\bC_n)$ subsequently. To emphasize on the specialized focus on the covariance matrix, we use the taxonomy of `coVariance filter' to denote graph filters implemented on covariance matrix as the graph.

In practice, the learning of coefficients \blue{$\bh$} in a supervised learning regime can be achieved via prevalent stochastic gradient descent that solves the optimization problem defined over a training set. The output of a covariance filter, i.e., $\bz = \bH(\hat\bC_n)\bx$ provides a data-driven representation built upon the covariance structure in~$\hat\bC_n$ through the filter coefficients $\bh$. In this scenario, the objective of the \emph{training} procedure is to determine the filter coefficients $\bh$ that provide the best possible fit of $\bz$ with some known ground truth (gauged through a pre-decided loss function, such as mean squared error \blue{(MSE)} for regression tasks or cross entropy loss for classification tasks)  for all inputs $\bx$ in the training set. The covariance structure $\hat\bC_n$ is estimated from the training dataset. The number of filter coefficients $K$ is a design choice, which can be determined through a hyperparameter optimization procedure~\cite{akiba2019optuna}. Figure~\ref{fig:pca_vf} illustrates the realization of a regression model with a coVariance filter and a readout function and its conceptual equivalence with a standard PCA-driven regression model.


%
%

Principles of GSP and GNNs have been the focus of recent tutorial treatments; see e.g.,~\cite{leus2023graph, ruiz2021graph}. Within this context, $h(\hat\lambda_i)$ is formally referred to as the \emph{frequency response} of the graph filter ${\bf H}(\cdot)$ and the coefficients $\bh$ are referred to as the \emph{filter taps}. The fundamental equivalence between PCA-driven information processing and coVariance filter-driven processing observation has wide implications, as it can be built upon to overcome the key limitations of PCA. Firstly, the polynomial implementation underlying the coVariance filter is computationally more efficient relative to conventional PCA approaches that rely on computationally expensive eigendecomposition operation on the covariance matrix. Consequently, the coVariance filter implementation overcomes instability of PCA-driven approaches stemming from unreliable estimates of the eigenspectrum (discussed in `Stability of VNNs' section). Moreover, the covariance filter $\bH(\cdot)$ is a \underline{scale-free} model as its parameters can be used to process a dataset of any arbitrary dimensionality \underline{without any changes to its architecture}. Specifically, the filter $\bH(\cdot)$ is contingent on the filter taps $\bh$ and not the dimensionality of the covariance matrix. Therefore, coVariance filter provides an analytic framework to theoretically analyze the transference across multiscale datasets~\cite{sihagJSTSP} (discussed in `Transferability of VNNs').    

\begin{figure}
    \centering
    \includegraphics[width=\linewidth]{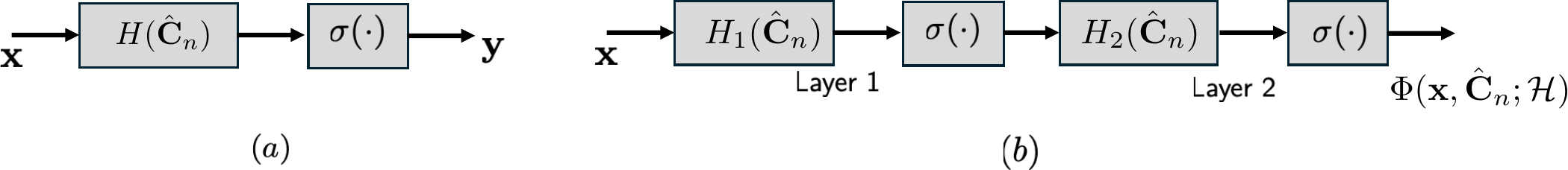}
    \caption{
    (a) Perceptron formed by a coVariance filter $\bH(\hat\bC_n)$ and a pointwise non-linearity function $\sigma(\cdot)$. (b) A 2-layer VNN.}
    \label{fig:vnn_arch}
\end{figure}

\noindent
{\bf From coVariance Filters to VNNs}
The representation capacity of a covariance filter is limited to learning \emph{linear} mappings over $\bx$ through $\bH(\hat\bC_n)$. The representation capacity herein can be extended to \emph{non-linear} mappings by nesting the covariance filter within a point-wise non-linear activation function; forming a coVariance perceptron. Specifically, for a given non-linear activation function $\sigma(\cdot)$, input $\bx$, a coVariance filter~${\bH( \hat\bC_n) = \sum_{k=0}^{K} h_k \hat\bC_n^k}$ and its corresponding filter coefficient set $\cH $, \blue{the coVariance perceptron is implemented as $\sigma(\bH( \hat\bC_n)\bx)$.}
Figure~\ref{fig:vnn_arch}a illustrates a coVariance perceptron. A coVariance neural network (VNN) can be constructed by cascading multiple layers of coVariance perceptrons (see Fig.~\ref{fig:vnn_arch} for a two-layer VNN). This observation is formalized next. 
\begin{remark}[Multi-layer VNN]\normalfont
 Consider an $L$-layer architecture formed by stacking $L$ coVariance perceptrons. In this scenario, we denote the coVariance filter in layer $\ell$ by~$\bH_{\ell}(\hat\bC_n)$ and its corresponding set of filter taps are given by $\cH_{\ell}$. For a given pointwise nonlinear activation function $\sigma(\cdot)$, the relationship between the input $\bx_{\ell-1}$ and the output $\bx_{\ell}$ for the coVariance perceptron in the $\ell$-th layer is given by
\begin{align}
    \bx_{\ell} = \sigma(\bH_{\ell}(\hat\bC_n)\bx_{\ell-1})\; \quad\text{for }\quad  \ell\in \{1,\dots,L\},
\end{align}
where $\bx_0$ is the input $\bx$. We refer to this $L$-layer architecture as an $L$-layer VNN. 
 \end{remark}
 The non-linear activation functions across different layers allow for non-linear transformations, thus, increasing the expressiveness of VNNs beyond linear mappings such as coVariance filters. In terms of architecture, VNNs are GNNs with covariance matrix as the graph shift operator~\cite{ruiz2021graph}. Similar to GNNs and other deep learning models~\cite{gama2020graphs, goodfellow2016deep}, the representation power of VNNs can be increased by incorporating multiple parallel inputs and outputs per layer enabled by filter banks at every layer. In general, we use the notation $\Phi({\bx};\hat\bC_n, \cH)$ to denote the representation learnt over the input $\bx$ by the VNN with $\cH$ as the set of all filter taps across all filters within the VNN. For a supervised learning objective with a training dataset $\{\bx_i, y_i\}_{i=1}^n$, the filter taps in $\cH$ are chosen to minimize the training loss, i.e.,
\begin{align}
    \cH_{\sf opt} = \min_{\cH} \frac{1}{n}\sum\limits_{i=1}^n J(\Phi(\bx_i;\hat\bC_n,\cH), y_i)\;,
\end{align}
where \blue{$J(\cdot)$ is some loss function that satisfies ${J(\Phi(\bx_i;\hat\bC_n,\cH), y_i) = 0}$} iff $\Phi(\bx_i;\hat\bC_n,\cH)= y_i$.  Since the covariance matrix $\hat\bC_n$ represents the covariance structure within the training set, we expect the mapping $\Phi(\bx;\hat\bC_n,\cH)$ to generalize well to data points $\bx$ that are outside the training set but come from the same distribution as the training set.

Existing theoretical GNN studies typically assume an abstract graph representation and overlook the unique aspects stemming from a data-driven graph model provided by an empirical (sample) covariance matrix~\cite{loukas2017close, joliffe1992principal}. Recent studies have undertaken rigorous theoretical analysis of VNNs~\cite{sihag2022covariance, sihagJSTSP, cavallo2024sparsecovarianceneuralnetworks, cavallo2024fair, cavallo2024spatiotemporal}; revealing unique theoretical and practical insights for deployment of GNNs in applications where covariance matrices inform the underlying graph structure. Furthermore, VNNs admit conceptual similarities with transformer architectures~\cite{vaswani2017attention}, where the self-attention mechanism in transformers can be interpreted as a generalized version of coVariance filters. This aspect has been discussed in `Self-Attention in Transformers versus CoVariance Filter in VNNs'. 

\begin{mdframed}[hidealllines=true,backgroundcolor=gray!20]
{\bf Self-Attention in Transformers versus CoVariance Filter in VNNs.}\\
Self-attention is the central information processing mechanism in the prevalent transformer-based deep learning architectures~\cite{vaswani2017attention}. Self-attention dynamically determines pairwise relevance based on feature representations, which motivates the broader interpretation of transformers as GNNs in prior studies~\cite{joshi2025transformers}. Interestingly, self-attention also admits a principled interpretation in terms of covariance, revealing conceptual connections to coVariance filters. Self-attention leverages the following learnable linear projections of data matrix $\bX \in \mathbb{R}^{m\times n}$:
\begin{align}
   {\bf Q} = \bX \bS_{\bQ},\quad \bK = \bX  \bS_{\bK},\quad \bV = \bX  \bS_{\bV}\;, 
\end{align}
where $\bS_{\bQ} \in \mathbb{R}^{n\times q}$, $\bS_{\bK} \in \mathbb{R}^{n\times q}$, and $\bS_{\bV} \in \mathbb{R}^{n\times v}$ are learnable matrices. The resulting matrix $\bQ$, $\bK$, and $\bV$ are referred to as query, key, and value matrices, respectively. By leveraging $\bQ$, $\bK$, and $\bV$, the attention matrix is computed as
\begin{align}\label{atteq}
    \bA = \mathrm{softmax}\!\left(\frac{\bQ\bK^\top}{\sqrt{q}}\right)
= \mathrm{softmax}\!\left(\frac{\bX \bS_{\bQ} \bS_{\bK}^\top \bX^\top}{\sqrt{q}}\right) =  \mathrm{softmax}\!\left(\frac{\bX \bM \bX^\top}{\sqrt{q}}\right)\;,
\end{align}
where $\bM = \bS_{\bQ} \bS_{\bK}^\top$ and the $\mathrm{softmax}$ function implements row-wise normalization. Therefore,~\eqref{atteq} reveals that evaluation of attention scores in $\bA$ is contingent upon the term $\bX \bM \bX^\top$, which can be viewed as a generalization (achieved via the learnable matrix $\bM$) of the standard covariance matrix employed in coVariance filters or VNNs. The output of the self-attention mechanism is given by
\begin{align}\label{otpt}
    \bO = \bA \bV\;.
\end{align}
The output $\bO$ takes the functional form of $\bO = f({\sf Cov}_{\bM}(\bX))\bX$, where $f$ is a non-linear normalization function and \blue{${\sf Cov}_{\bM}(\bX) = \bX\bM\bX^\top$}. On the other hand, the output of the coVariance filter in~\eqref{gft3} over the complete dataset admits the form $\bZ = g\Big(\frac{1}{n-1}{\sf Cov}_{\bI_{m}}(\bX)\Big)\bX$, where $g$ is a linear function, $\bI_{m}$ is the identity matrix of size $m\times m$ and therefore, \blue{$\frac{1}{n-1}{\sf Cov}_{\bI_{m}}(\bX)$ is the sample covariance matrix (as ${\sf Cov}_{\bM}(\bX)$ reduces to $(n-1)\hat\bC_n$ when $\bM = \bI_m$)}. Thus, the self-attention mechanism processes a generalized form of covariance via weights in $\bM$ to \blue{produce an} output. When $\bM$ is set to \blue{the identity} matrix, self-attention mechanism yields a coVariance filter with non-linear normalization. Similar arguments extend to yield the parallels between multi-head attention and a filter bank of coVariance filters.  

Our discussion above reveals covariance as the common ground between the self-attention mechanism and coVariance filters. The superiority of attention-driven transformer architectures to other alternative learning models largely stems from their superior representation capacity, which comes at the cost of needing larger datasets for achieving similar performance as architectures with inbuilt specialized inductive biases (for instance, vision transformer versus a convolutional neural network~\cite{raghu2021vision}). Large-scale datasets may not be feasible in various real-world applications (for instance, in biomedical applications that require considerable resources to build huge datasets). For such scenarios with limited data, VNNs provide an effective deep learning solution that is intricately connected to the fundamental principles of transformers.

\end{mdframed}

\section*{Stability of VNNs}
We reiterate from `Challenges to Learning with PCA' that the reproducibility of inference outcomes through PCA-driven approaches is challenged by the statistical uncertainty in the estimation of principal components of $\hat\bC_n$. Specifically, the eigenvalues $\{\hat\lambda_1,\dots, \hat\lambda_m\}$ of $\hat\bC_n$ are likely to be perturbed relative to the eigenvalues $\{\lambda_1,\dots, \lambda_m\}$ of $\bC$. Hence, for close eigenvalues of $\bC$, the corresponding estimates in $\hat\bC_n$ may not maintain consistent ordering with a high likelihood (see~\eqref{eiggap}). Such considerations are relevant to learning with VNNs or PCA because we typically operate with $\hat\bC_n$ rather than $\bC$ in practical applications. Moreover, a traditional PCA-driven approach is highly vulnerable to irreproducibility when $\hat\bC_n$ is perturbed relative to its estimate from the given training dataset (for instance, through a re-estimation of $\hat\bC_n$ with more samples or in an independently collected dataset; see Fig.~\ref{fig:spm_stab}). In the context set above, the problem of establishing the stability of VNNs can be informally stated as follows.

\noindent
{\bf Stability problem (Informal).} \textit{Given an input data sample $\bx$, we aim to theoretically characterize the divergence between the representations $\Phi(\bx; \hat\bC_n, \cH)$ and $\Phi(\bx; \bC, \cH)$. If $\Phi(\bx; \hat\bC_n, \cH)$ and $\Phi(\bx; \bC, \cH)$ converge meaningfully (for instance, with \blue{an} increase in $n$), we can conclude that the VNN learns representations that are robust to the stochastic perturbations in $\hat\bC_n$ relative to $\bC$. }

CoVariance filter and VNN are discussed with respect to the sample covariance matrix $\hat\bC_n$ estimated from the training dataset in~`From CoVariance Filters to VNNs'. In this context, the notion of stability of VNNs (or coVariance filters) is formalized through the impact of statistical uncertainty induced in the sample covariance matrix with respect to the ensemble covariance matrix on the coVariance filter output. To this end, the primary goal is to investigate the divergence $\|\Phi(\bx;\hat\bC_n, \cH) - \Phi(\bx; \bC, \cH)\|$. If $\|\Phi(\bx;\hat\bC_n, \cH) - \Phi(\bx; \bC, \cH)\|$ is controlled in some sense, we expect the learnable parameters $\cH$ to meaningfully exploit the covariance structure in $\hat\bC_n$, such that, a VNN with parameters $\cH$ provides performance robust to the selection of $\hat\bC_n$ and not overfit on the given specific instance of the training set.  Thus, establishing stability within this context provides the critical theoretical guarantees on the robustness of representations learned by a VNN to stochastic perturbations in $\hat\bC_n$ relative to $\bC$, and consequently, the reproducibility of inference outcomes by coVariance filters or VNNs. Such theoretical guarantees are often missing from PCA-driven learning approaches. Theorem~\ref{filterstab} characterizes the stability of coVariance filters and VNNs.

\begin{theorem}[Stability of coVariance Filters and VNNs~\cite{sihag2022covariance}]\label{filterstab}
Consider a random vector $\bx \in \mR^{m\times 1}$, such that, \blue{$\|\bx\|_2\leq 1$}, and its corresponding ensemble covariance matrix $\bC = \mathbb{E}[(\bx - \mE[\bx])(\bx - \mE[\bx])^{\sf T}]$. For a sample covariance matrix $\hat\bC_n$ formed using $n$ i.i.d instances of $\bx$ \blue{and a coVariance filter with the filter taps $\bh$ and the frequency response $h(\cdot)$}, if the frequency response satisfies {$|h(\lambda_i)- h(\lambda_j)|\leq \varsigma|\lambda_i - \lambda_j|,  \forall i\neq j$ for an appropriate $\varsigma>0$}, the following holds with \blue{a probability at least $1 - n^{-2\varepsilon} - 2 \kappa m/n$}:
\begin{align}\label{filterstab_rslt}
    \blue{\left\lVert \bH(\hat\bC_n) - \bH(\bC)\right\rVert_{\sf op} \leq \frac{\varsigma}{n^{1/2-\varepsilon}} {\cal O}\left(\sqrt{m} + \frac{\|\bC\|\sqrt{\log(nm)}}{\mu n}\right) = \alpha_n\;,}
\end{align}
for any $\varepsilon \in(0,1/2)$ and a constant $\kappa>0$ that depends on the data distribution.  Further, \blue{for a VNN $\Phi(\cdot;\cdot,{\cal H})$ consisting of filter banks with $F$ filters per layer and $L$ number of layers}, if the pointwise non-linearity $\sigma(\cdot)$ satisfies $|\sigma(a) - \sigma(b)|\leq |a-b|$, then, 
\begin{align}\label{vnnbound}
    \|\Phi(\bx;\hat\bC_n, \cH)-\Phi(\bx;\bC,\cH)\| \leq LF^{L-1} \alpha_n\;.
\end{align}
\end{theorem}
The right-hand side term in~\eqref{filterstab_rslt} and the condition $|h(\lambda_i)- h(\lambda_j)|\leq \varsigma|\lambda_i - \lambda_j|$ are products of the analysis of the finite sample size effect-driven perturbations in $\hat\bC_n$ and its eigenvectors with respect to that in $\bC$. From Theorem~\ref{filterstab}, the divergence $ \|\bH(\hat\bC_n) - \bH(\bC)\|$ decays with the number of samples $n$ at least at the rate of $1/n^{\frac{1}{2} - \varepsilon}$. Thus, the representations learned by VNN with $\hat\bC_n$ as the covariance matrix converges with that learned by the VNN with $\bC$ and same parameters $\cH$. 

The cost for the stability in~\eqref{filterstab_rslt} is also apparent from Theorem~\ref{filterstab}. Specifically, for a given eigenvalue $\lambda_i$, the response of the filter for any eigenvalue $\lambda_j, j\neq i$ becomes closer to $h(\lambda_i)$ with \blue{a} decrease in $|\lambda_i - \lambda_j|$. Recall from the perturbation theory of eigenvectors and eigenvalues of sample covariance matrices that the sample-based estimates of the eigenspaces corresponding to eigenvalues $\lambda_i$ and $\lambda_j$ become harder to distinguish as $|\lambda_i - \lambda_j|$ decreases~\cite{loukas2017close}. Therefore, the coVariance filter that satisfies~\eqref{filterstab_rslt} sacrifices discriminability between close eigenvalues to preserve its stability with respect to the statistical uncertainty inherent in the sample covariance matrix. Specifically, the variation between frequency responses of the coVariance filter for any two eigenvalues $\lambda_i$ and $\lambda_j$ in Fig.~\ref{fig:pca_vf}, i.e., $|h(\lambda_i)-h(\lambda_j)|$ is governed by the difference $|\lambda_i - \lambda_j|$.

In contrast to coVariance filters, the VNNs incorporate non-linear activation functions within their information processing structure. Similar to GNNs~\cite{gama2020graphs}, we expect the VNNs to retain the discriminability between close eigenvalues due to the demodulation impact of non-linear activation function. Case Study 1 provides an empirical study that corroborates the stability of VNN outcomes on a linear regression task while demonstrating the potential stability of PCA-driven models. 

\subsection*{Sparse VNNs}
The sample covariance matrix $\mathbf{\hat{C}}_n$ often contains spurious correlations and finite-sample estimation errors that produce inaccurate principal directions and result in a dense estimate.
This hinders the estimation quality if the true covariance is sparse, and is detrimental for computational efficiency.
While sparse PCA is a traditional remedy for improving estimation quality and reducing computation~\cite{deshpande2016sparse, bickel2008covariance}, it suffers from the same instabilities as standard PCA.
To address these limitations,
Sparse VNNs (SVNNs)~\cite{cavallo2024sparsecovarianceneuralnetworks} integrate sparse covariance estimators directly into the VNN framework.
If the true covariance is known or expected to be sparse, SVNNs employ hard thresholding, i.e., a function $\eta(\cdot)$ that acts on each \blue{entry $\hat c_{ij}$ of the covariance matrix} as
\begin{equation}
    \eta(\mathbf{\hat{C}}_n)_{ij} = 
\begin{cases} 
\hat c_{ij} & \text{if } |\hat c_{ij}| > \frac{\tau}{\sqrt{n}}, \\
0 & \text{otherwise}.
\end{cases}
\end{equation}
That is, hard thresholding sets to zero the elements below the threshold. This covariance estimator admits the following stability analysis.

\begin{theorem}[Stability of deterministic sparse covariance filters~\cite{cavallo2024sparsecovarianceneuralnetworks}]\label{sparsefilterstab}
Consider a random vector $\bx \in \mR^{m\times 1}$, such that, $|\bx|\leq 1$, and its corresponding ensemble covariance matrix $\bC = \mathbb{E}[(\bx - \mE[\bx])(\bx - \mE[\bx])^{\sf T}]$ with at most $c_0$ non-zero elements per row. Consider also a sample covariance matrix $\hat\bC_n$ formed using $n$ i.i.d instances of $\bx$, and let the frequency response satisfy {$|h(\lambda_i)- h(\lambda_j)|\leq \varsigma|\lambda_i - \lambda_j|$ for an appropriate $\varsigma>0$}. 
Consider the application to the sample covariance of a hard-thresholding function $\eta(\cdot)$ with threshold $\tau = M'\sqrt{\log(m)}$ and $M'$ a large enough constant.
Let the eigenvalues of the true covariance $\{\lambda_i\}_{i=1}^{N}$ be all distinct and strictly positive. Then, the following holds with high probability:
\begin{align}\label{sparsefiltersbound}
    \|\mathbf{H}(\eta(\mathbf{\hat C}_n))-\mathbf{H}(\mathbf{C})\| \leq 
    \mathcal{O}\left( \frac{\varsigma c_0 m\sqrt{\log{m}}}{n^{1/2}} \right).
\end{align}    
\end{theorem}
Theorem~\ref{sparsefilterstab} shows that the main takeaways of the stability analysis for covariance filters in Theorem~\ref{filterstab} also apply to thresholded covariances. \blue{In particular, the bound decreases with the number of samples $n$, increases with the sample dimensions $m$ and with the constant $\varsigma$. Differently from standard covariance filters, however, the bound in~\eqref{sparsefiltersbound} depends on the number of non-zero elements per row $c_0$ instead of the spectral norm of the covariance $\|\bC\|$, which tightens it for sparse covariances. This occurs because the thresholding operation bounds the spectral error via the maximum absolute row sum, confining the accumulation of estimation noise to the $c_0$ true non-zero entries. Furthermore, the threshold parameters $\tau$ and $M'$ do not appear in the final bound because they are not independent constants. Rather, they are chosen to match the maximum expected statistical noise of individual entries, meaning their effect is absorbed into the probability bound that guarantees the recovery of the true zeros (cf. \cite{bickel2008covariance}).} These considerations extend to sparse VNNs using~\eqref{vnnbound}.

However, the assumption of a sparse true covariance might not hold in practice, limiting the applicability and the validity of the stability results for soft thresholding. 
Therefore, SVNNs also consider stochastic sparsification, an approach where covariance values $\hat{c}_{ij}$ are dropped with probabilities $p_{ij}$ that can be specified based on the application needs. For example, probabilities can be set to achieve a desired sparsification level, which guarantees a desired computational efficiency improvement, but might lose relevant covariance information. This results in the following stability result.

\begin{theorem}[Stability of deterministic sparse covariance filters~\cite{cavallo2024sparsecovarianceneuralnetworks}]
Consider a random vector $\bx \in \mR^{m\times 1}$, such that, $|\bx|\leq 1$, and its corresponding ensemble covariance matrix $\bC = \mathbb{E}[(\bx - \mE[\bx])(\bx - \mE[\bx])^{\sf T}]$ with at most $c_0$ non-zero elements per row. Consider also a sample covariance matrix $\hat\bC_n$ formed using $n$ i.i.d instances of $\bx$, and let the frequency response be generalized integral Lipschitz with constant $\varsigma$ (see~\cite{cavallo2024sparsecovarianceneuralnetworks} for a formal definition).
Consider applying to the sample covariance a  stochastic thresholding function $\eta(\cdot)$ with probabilities $p_{ij}$. Then, the following holds with high probability
\begin{equation}
     \mathbb{E}\left[\left\lVert \bH(\eta(\hat\bC_n)) - \bH(\bC)\right\rVert ^2 \right] \leq \alpha_n^2 + m \varsigma^2 \sum_{i=1}^m\sum_{j=1}^m \hat{c}_{ij}^2 (1-p_{ij}),
\end{equation}
where $\alpha_n$ is defined in \eqref{filterstab_rslt}.
\end{theorem}
In this case, SVNNs' stability is affected by two sources of error: the finite-sample estimation $\alpha_n$, which decreases with the number of samples, and the sparsification, which depends on the magnitude and quantity of covariance elements dropped.
This indicates that removing large covariances $\hat{c}_{ij}^2$ might result in lower stability as relevant information is lost, which identifies a sparsity-stability tradeoff that can be controlled by the dropping probabilities $p_{ij}$. Moreover, this bound extends to complete sparse VNN architectures by applying~\eqref{vnnbound}.

\begin{figure}[t]
  \centering
  \includegraphics[scale=0.55]{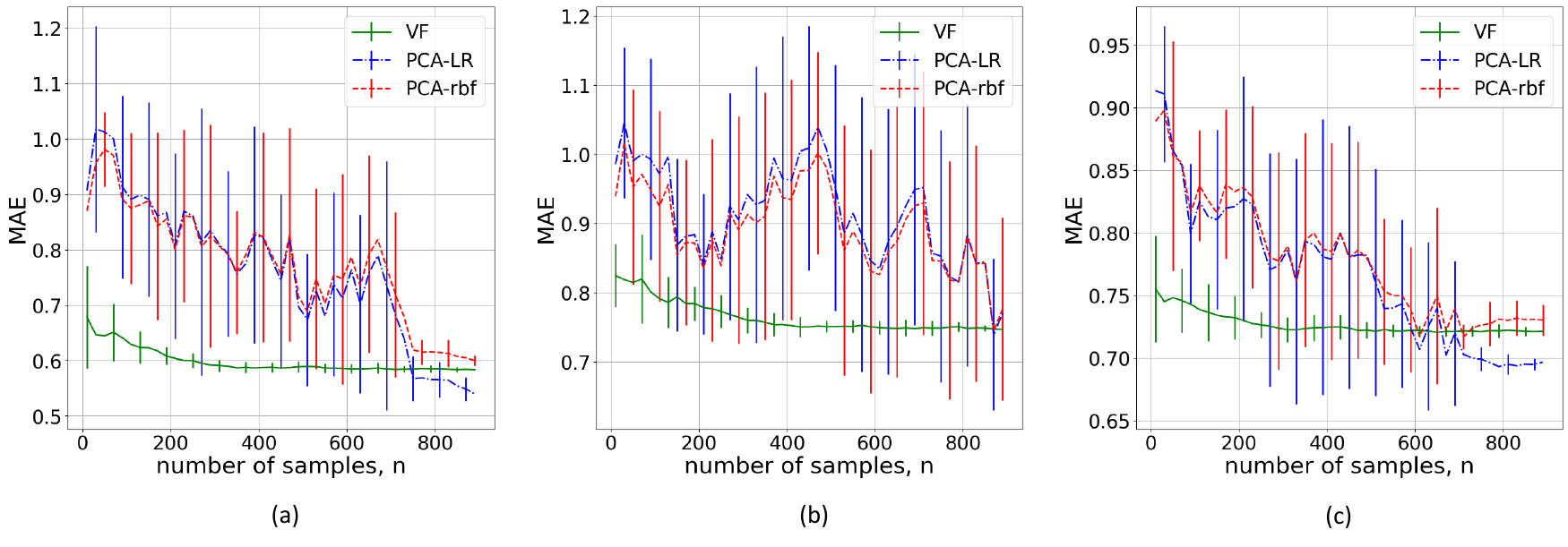}
   \caption{\blue{Stability of coVariance filter (VF)-based architectures on three linear regression problems. (a), (b), and (c) illustrate the variations in MAE performances over three distinct datasets with respect to perturbations in the sample covariance matrix used during training, i.e., $\hat\bC_{900}$.}}
   \label{LR07}
\end{figure}

\begin{mdframed}[hidealllines=true,backgroundcolor=gray!20]
\blue{{\bf Case Study 1. Stability of coVariance filters.}}\\ 
\blue{In this case study, we investigate the stability of coVariance filter (VF)-based architectures relative to PCA-driven models on three regression tasks. }

 \noindent
\blue{{\bf Datasets.} For this purpose, we generate random linear regression problems using the Python routine \texttt{sklearn.datasets.make\_regression}. For all datasets, we set the dimensionality of the input data to be $m=20$, the dimension of the response to be $1$, the number of samples as $n=5000$, the effective rank of the input dataset to be $10$. The standard deviation of the Gaussian noise applied to the output is set to $3$, $5$, and $7$ to generate three distinct datasets.} 

\noindent
\blue{{\bf Experiments.} Our primary aim is to compare the robustness of the outcomes of VF-driven model and PCA-driven regression models to perturbations in the sample covariance matrix. Additionally, we also check whether the trained VFs satisfied the condition of Lipschitz continuity within Theorem~\ref{filterstab}. For each setting, the dataset was split into a $90/10$ train/test split, and the sample covariance matrix $\hat\bC_{900}$ was generated from the respective training set. The VF-driven model consisted of a single filter bank of 10 VFs (operating in parallel) that processed the input data. The outputs at all filters were aggregated using unweighted mean function to yield the output. This model is denoted as `VF' in the results in Fig.~\ref{LR07}. The PCA-regression pipeline consisted of two steps: i) we first identified the principal components using the eigendecomposition of $\hat\bC_{900}$; and then, ii) to maintain consistency with VNN, transformed the training set used for VNN training to fit to the corresponding response data using a regression model. Regression in the PCA-regression pipelines was implemented with ‘linear’ and radial basis function (‘rbf’) kernels. PCA-regression with linear kernel in the regression model is referred to as PCA-LR and that
with ‘rbf’ kernel in the regression model is referred to as PCA-rbf. The optimal number of principal components in the PCA-regression pipeline was selected through a 10-fold cross-validation procedure on the training set, repeated 5 times. Using different permutations of the training set, we obtained $10$ nominal models for VF, \blue{${\sf PCA-LR}$}, and ${\sf PCA-rbf}$.}

\noindent
\blue{{\bf Results.} Figures~\ref{LR07}a-c plot the variations in the mean absolute error (MAE) performances of the nominal models on the test set with respect to perturbations in the sample covariance matrix, with the last data point corresponding to $n=900$, i.e., $\hat\bC_{900}$. When $\hat\bC_{900}$ was replaced with $\hat\bC_{n'}$ for any $n'\in[10,899]$, our experiments show that VNN performance is stable, but the performances of {\sf PCA-LR} and {\sf PCA-rbf} models were highly stochastic. Theorem~\ref{filterstab} guaranteed the stability of a VF, provided its frequency response was Lipschitz continuous. In this context, our experiments revealed that the VFs across the three settings were indeed Lipschitz continuous, with the estimated Lipschitz constant being $19\pm 2.12$, $19.09\pm3.37$, and $10.94\pm 1.97$ for VFs trained on dataset with the standard deviation of noise $3$, $5$, and $7$, respectively. We also remark that the Lipschitz continuity and the corresponding constants could be explicitly enforced and controlled by adding appropriate penalty terms to the loss function during training~\cite{gama2020stability, ruiz2021graph}. }


\end{mdframed}

\begin{figure}[t]
  \centering
  \includegraphics[scale=0.55]{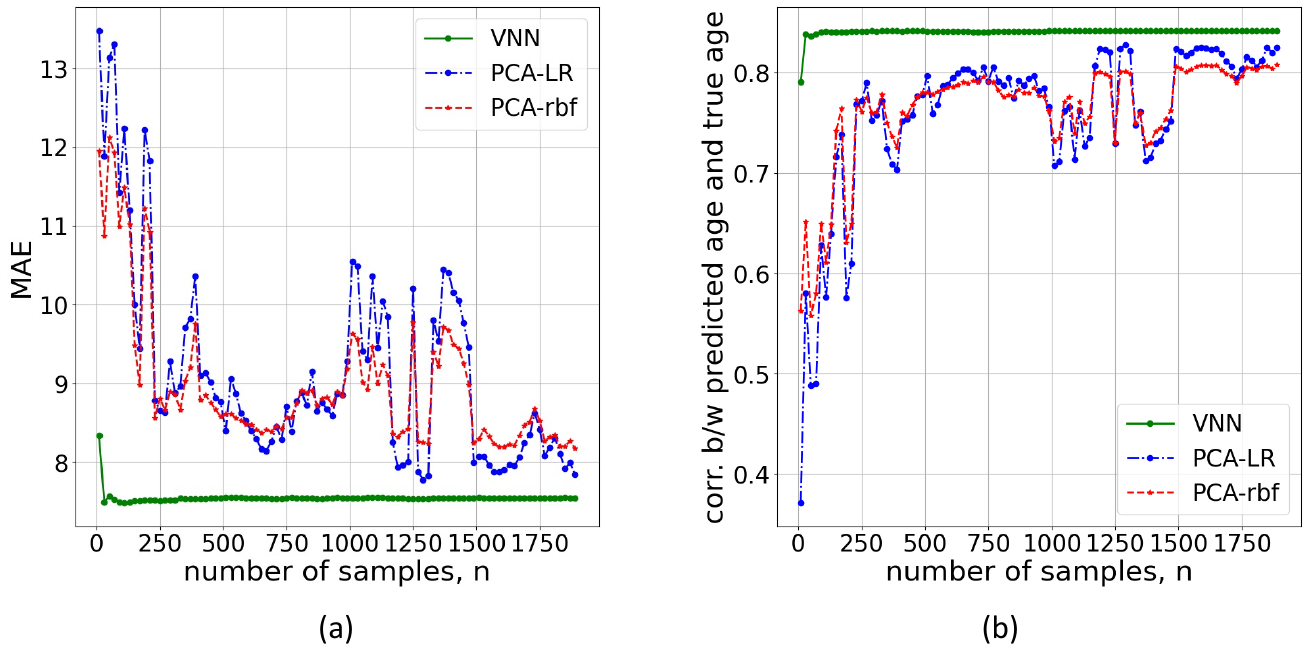}
   \caption{\blue{Stability of VNN outcomes on the age prediction task. (a) and (b) illustrate the variations in MAE and Pearson's correlation between the true age and predicted age with respect to perturbations in the sample covariance matrix used during the training of VNN, {\sf PCA-LR} and {\sf PCA-rbf} models.}}
   \label{LR08}
\end{figure}

\begin{mdframed}[hidealllines=true,backgroundcolor=gray!20]
\blue{{\bf Case Study 2. Stability of VNNs.}}\\ 
\blue{In this case study, we study the stability of the outcomes of VNN models on a real world dataset. The dataset consists of `cortical thickness' features derived from structural magnetic resonance} \blue{imaging (MRI) for each individual in a healthy population. A covariance matrix of cortical thickness features (referred to as anatomical covariance matrix~\cite{evans2013networks}), captures how thickness measurements from different brain regions vary together across individuals, with off-diagonal elements representing inter-regional covariances between thickness measures. Moreover, cortical thickness features evolve across the lifespan and hence, are meaningful representations of healthy aging phenomenon. In this case study, we focus on predicting age from cortical thickness features by leveraging the anatomical covariance matrix in the models deployed for this task. PCA is a widely used approach in this context because it exploits the dominant directions of cortical thickness variation from anatomical covariance information for downstream inference tasks~\cite{mcwhinney2024principal,cao2023cortical}.} 

\noindent
\blue{{\bf Datasets.} We used the cortical thickness features derived from T1-weighted (T1w) MRI images collected from 3.0 Tesla MRI scanners for several publicly available datasets constituted by $2147$ healthy individuals who were 18 years or older. These datasets include: (a) Cambridge Centre for Ageing and Neuroscience (CamCAN) dataset~\cite{shafto2014cambridge}; (b) Dallas Lifespan Brain Study (DLBS) \href{https://fcon\textunderscore 1000.projects.nitrc.org/indi/retro/dlbs.html}{https://fcon\textunderscore 1000.projects.nitrc.org/indi/retro/dlbs.html}; (c) IXI dataset (\href{https://brain-development.org/ixi-dataset/}{https://brain-development.org/ixi-dataset/}; and enhanced Nathan Kline Institute-Rockland Sample (eNKI)~\cite{nooner2012nki}. The demographics of these datasets are summarized in Table~\ref{table_demo}. The T1w MRI images in these datasets were preprocessed via the open-source CAT12 pipeline~\cite{gaser2022cat} to yield cortical thickness measures for each individual. For every individual, we obtained cortical thickness measures curated according to 100 parcellations version of Schaefer's brain atlas~\cite{schaefer2018local}.} 

\noindent
\blue{{\bf Architecture and training.} VNN and PCA-regression models were trained to predict the chronological age using $100$ cortical thickness features across the cortex for the population of healthy individuals, described in Table~\ref{table_demo}. The VNN architecture consisted of $2$-layers, with $2$ filter taps each in the two layers. The width of the network was set to $13$. The readout function at the final layer was a learnable linear layer that mapped the outputs at the final layer of the VNN to the scalar estimate of chronological age. The training procedure was based on optimizing the \blue{MSE} loss function. The learning rate for the Adam optimizer was $0.01$. The model was trained on $90\%$ of the dataset and tested on the remaining $10\%$ to validate that it could infer information about chronological age on an unseen healthy population. The $90/10$ split of the dataset was determined randomly. The model operated on the anatomical covariance matrix of size $100\times 100$ estimated from the $100$ cortical thickness features of the $90\%$ split and normalized, such that its maximum eigenvalue was $1$. Furthermore, the cortical thickness features were z-score normalized across the $90\%$ split and this normalization was applied to the $10\%$ validation cohort. VNN achieved MAE of $7.54$ years with correlation between the ground truth and VNN estimates of chronological age being $0.841$ on the test set. PCA-regression models achieved similar performances on the test set: {\sf PCA-LR} achieved MAE of $7.64$ years and Pearson's correlation of $0.832$ on the test set, {\sf PCA-rbf} achieved MAE of $8.03$ years and Pearson's correlation of $0.813$ on the test set. The age prediction performances of VNNs and PCA-regression models are consistent with those reported in the literature that used cortical thickness datasets from similar populations~\cite{madan2018predicting, guan2024brain}. Our objective here was to evaluate the stability of the aforementioned performances of VNN and PCA-regression models as the sample covariance matrix used during training was replaced with re-estimated covariance matrices from smaller number of data samples while the learned parameters were frozen.} 	

\noindent
\blue{{\bf Results.} Figures~\ref{LR08}a-b plot the variations in the MAE performances and Pearson's correlation between the ground truth and estimated age on the test set with respect to perturbations in the sample covariance matrix, with the last data point corresponding to $n=1932$, i.e., $\hat\bC_{1932}$. VNN performance was observed to be highly stable, but the performances of {\sf PCA-LR} and {\sf PCA-rbf} models were highly stochastic (Figs.~\ref{LR08}a-b). These observations were consistent the theoretical guarantees on the stability of VNNs in Theorem~\ref{filterstab}, while no such guarantees exist for PCA-driven models. }


\end{mdframed}
\begin{figure}
    \centering
    \includegraphics[width=\linewidth]{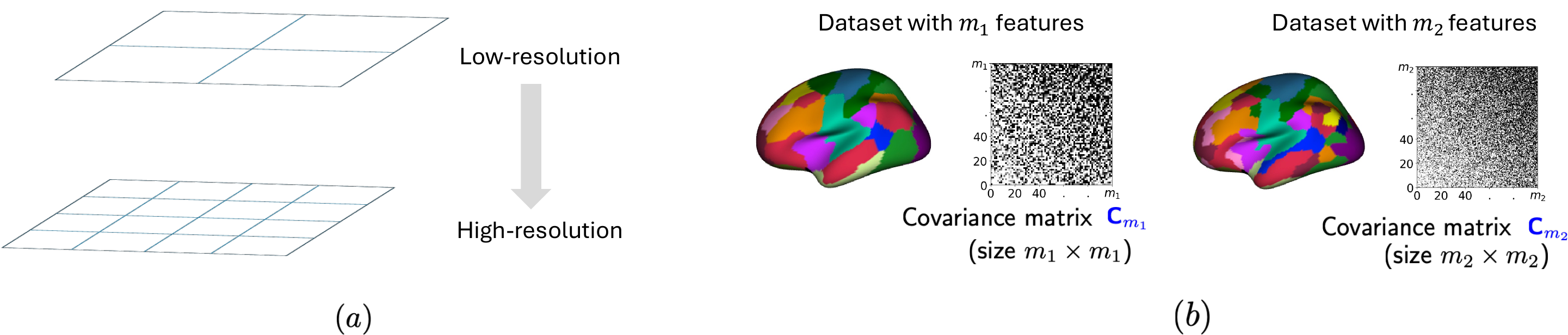}
    \caption{(a) Abstract representation of a multiscale organization of a dataset. (b) Multiscale neuroimaging dataset illustrating partition of the brain surface at $m_1$ or $m_2$ resolutions with corresponding true covariance matrices $\bC_{m_1}$ and $\bC_{m_2}$. }
    \label{mscale}
\end{figure}

\begin{table}[h]
\setlength{\tabcolsep}{3.5pt}
\caption{Demographics for datasets used in Case Studies 2 and 3.}

\centering
\renewcommand{\arraystretch}{1}
\begin{threeparttable}
{\begin{tabular}{|c| c| c| c|c|}
\hline
\multirow{2}{*}{Dataset} & \multirow{2}{*}{n} & \multirow{2}{*}{Sex (m/f)} & \multicolumn{2}{c|}{Age}\\ 
\cline{4-5}
& & & range & mean $\pm$ s.d. \\
\hline
 CamCAN & 652 & 322/330 & 18.5-88.92 & 54.77 $\!\pm\!$ 18.6 \\
\hline
 DLBS & 315 & 117/198 & 20.57 -89.11& 54.62 $\pm$20.09 \\
 \hline
 IXI & 182& 88/94 &20.16-81.94 & 47.44$\pm$16.7 \\
 \hline
 eNKI & 998& 347/650\tnote{*} &18-85 &47.35$\pm$17.71 \\
\hline
 {\bf Total} & 2147 & 874/1272\tnote{*} &18-89.11 &50.68$\pm$18.62 \\
\hline
\end{tabular}}
\begin{tablenotes}\footnotesize
\item[*] Sex information missing for 1 individual.
\end{tablenotes}
\end{threeparttable}
\label{table_demo}
\end{table}

\section*{Transferability of VNNs}
In the multiscale setting with spatial resolutions $m_1$ and $m_2$, the dataset with $m_1$ features and $n$ data samples is denoted by \blue{${\bf X}_{m_1} \in \mathbb{R}^{m_1\times n}$}  and that with $m_2$ features and $n$ data samples is denoted by \blue{${\bf X}_{m_2} \in \mathbb{R}^{m_2\times n}$}. Both datasets are assumed to represent the same information at distinct resolutions. See Fig.~\ref{mscale}a for an abstract representation of the multiscale setting considered here. Figure~\ref{mscale}b illustrates the multiscale organization in the setting of a neuroimaging dataset curated according to two distinct resolutions of a brain atlas. The corresponding true covariance matrices for the datasets with $m_1$ and $m_2$ features are denoted by $ {\bf C}_{m_1}$ and  $ {\bf C}_{m_2}$, respectively. For conciseness in exposition and avoiding repetition of arguments, we have focused our discussions in this section on the true covariance matrices as opposed to sample covariance matrices in the previous sections. The problem statement of the transference of VNNs across multiscale datasets is informally stated next.

\noindent \textit{{\bf Transference problem  (Informal)}. Given a data point ${\bf x}_{m_1}$ from the dataset with $m_1$ features and associated covariance matrix ${\bf C}_{m_1}$, and another data point ${\bf x}_{m_2}$ from the dataset with $m_2$ features and associated covariance matrix ${\bf C}_{m_2}$, we aim to characterize the operator {distance} between the representations $\Phi({\bf x}_{m_1};{\bf C}_{m_1},{\cal H})$ and $\Phi({\bf x}_{m_2};{\bf C}_{m_2},{\cal H})$. If $\Phi({\bf x}_{m_1};{\bf C}_{m_1},{\cal H})$ and $\Phi({\bf x}_{m_2};{\bf C}_{m_2},{\cal H})$ converge in some sense, we can conclude that the VNN with parameters ${\cal H}$ is transferable between two datasets consisting of $m_1$ and $m_2$ features.}

The above problem involves comparing the representations $\Phi({\bf x}_{m_1};{\bf C}_{m_1},{\cal H})$ and $\Phi({\bf x}_{m_2};{\bf C}_{m_2},{\cal H})$, which are of different dimensionalities and hence, this comparison is not natural. This comparison is facilitated theoretically by the mapping of the datasets and associated representations to the same continuous domain~\cite{sihagJSTSP}. As an example of a scenario where such a mapping could be valid, consider two neuroimaging datasets curated according to different brain atlases (and hence, having different number of features) but capturing brain activity across the whole brain. In this context, the scope of information for both datasets is limited to the same brain surface. Since the datasets with $m_1$ and $m_2$ features can be imagined as discrete samples of the same continuum (brain surface), we can compare them mathematically by mapping them to the same continuous domain.

\begin{figure}
    \centering
    \includegraphics[width=0.75\linewidth]{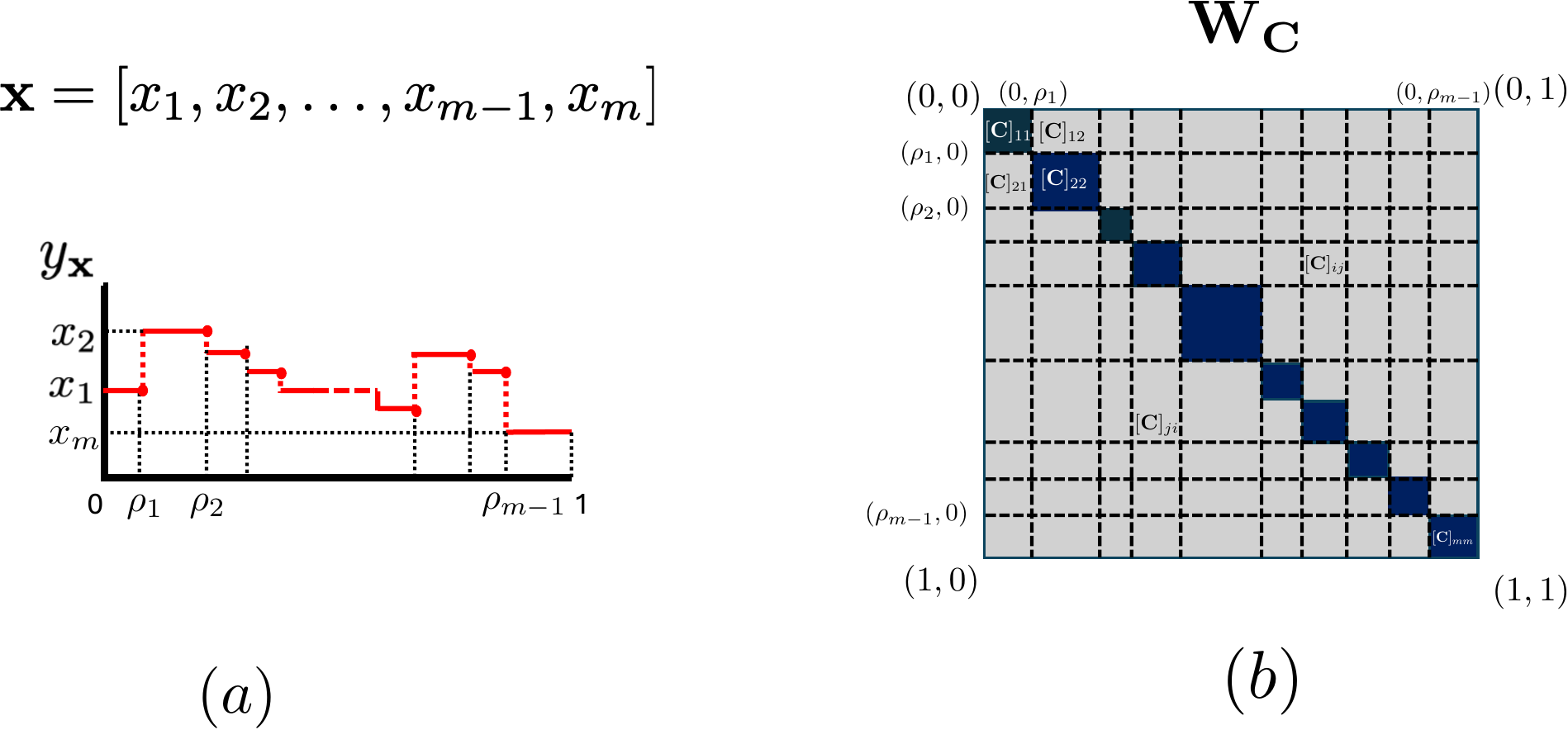}
    \caption{ Continuous approximations of (a) $m$-dimensional discrete data $\bx$ in the interval $[0,1]$ and (b) an $m\times m$ covariance matrix in the space $[0,1]^2$. }
    \label{fig:contapprox}
\end{figure}

In previous works~\cite{sihagJSTSP}, the following mapping of data to the continuous domain has been considered: Given an $m$-dimensional vector ${\bf x} = [x_1,\dots,x_m]$, we can define a continuous representation of ${\bf x}$ as a function $y_{{\bf x}}: [0,1] \mapsto \mathbb{R}$, such that, $y_{{\bf x}}(u) = x_i$ for $u\in {\cal U}_i$, where ${\cal U}_i$ is a predefined interval associated with the $i$-th element of ${\bf x}$, and proportional to the relative \textit{variance} of a feature. Similarly, we can map a covariance matrix ${\bf C}_m$ to a compact set $[0,1]^2$ using the mapping ${\bf W}_{{\bf C}_m}: [0,1]^2 \mapsto \mathbb{R}$, where we have ${\bf W}_{{\bf C}_m}(u,v) = [{\bf C}_m]_{ij}$ for $u\in {\cal U}_i$ and $v\in {\cal U}_j$. A pictorial illustration of $y_{{\bf x}}$ and ${\bf W}_{{\bf C}}$ for covariance matrix ${\bf C}$ is included in Fig.~\ref{fig:contapprox}.

The consequence of having a continuous approximation is that, given the setting with multiscale data points ${\bf x}_{m_1}$ and ${\bf x}_{m_2}$, the closeness of continuous representations $y_{{\bf x}_{m_1}}$ and $y_{{\bf x}_{m_2}}$  can be used as a metric to assess the similarity between them. This observation also extends to the theoretical comparisons between matrices ${\bf C}_{m_1}$ and ${\bf C}_{m_2}$ or representations $\Phi({\bf x}_{m_1};{\bf C}_{m_1},{\cal H})$ and $\Phi({\bf x}_{m_2};{\bf C}_{m_2},{\cal H})$. Hence, the transference problem for VNNs can be formalized mathematically as follows. 

\noindent
{\bf Transference problem with continuous approximations:} Consider two representations $\Phi({\bf x}_{m_1};{\bf C}_{m_1},{\cal H})$ and $\Phi({\bf x}_{m_2};{\bf C}_{m_2},{\cal H})$ derived by the same covariance filter coefficients $\cH$ on datasets with $m_1$ and $m_2$ features, respectively. If we have the following conditions: (a) the continuous approximations of inputs ${\bf x}_{m_1}$ and ${\bf x}_{m_2}$ are close, i.e., $\|y_{{\bf x}_{m_1}} - y_{{\bf x}_{m_2}}\|$ is bounded; and (b) the continuous approximations of covariance matrices ${\bf C}_{m_1}$ and ${\bf C}_{m_2}$ are close, i.e., ${\|{\bf W}_{{\bf C}_{m_1}} - {\bf W}_{{\bf C}_{m_2}}\|}$ is bounded; we aim to characterize the closeness between the continuous approximations of representations $\Phi({\bf x}_{m_1};{\bf C}_{m_1},{\cal H})$ and $\Phi({\bf x}_{m_2};{\bf C}_{m_2},{\cal H})$, i.e., find $\vartheta >0$, such that, ${\|y_{\Phi({\bf x}_{m_1};{\bf C}_{m_1},{\cal H})}- y_{\Phi({\bf x}_{m_2};{\bf C}_{m_2},{\cal H})}\|_2\leq \vartheta.}$

For conciseness, we subsequently use the notation $y_{m_1}$ (and $y_{m_1}$) to denote $y_{\Phi({\bf x}_{m_1};{\bf C}_{m_1},{\cal H})}$ (and $y_{\Phi({\bf x}_{m_2};{\bf C}_{m_2},{\cal H})}$). Establishing the transference problem as above provides a novel perspective to the domain of transfer learning in multiscale settings, where VNNs provide novel insights regarding the interplay between the redundancy in information across multiple scales and the quality of transference of VNNs. Such theoretical insights are simply infeasible with prevalent deep learning networks in this context~\cite{wang2023trend, wang2024deep}. \blue{Transference of VNNs is quantified in terms of the metric $\|y_{m_1} - y_{m_1}\|_2 = \Big(\int_{0}^{1} (y_{m_1} (r) - y_{m_2}(r))^2{\sf d}r\Big)^{\frac{1}{2}} $ in Theorem~\ref{transferthm}, where $\|y_{m_1} - y_{m_1}\|_2$ is a measure of dissimilarity between the continuous representations $y_{m_1}$ and $y_{m_2}$.}

\begin{figure}
    \centering
    \includegraphics[width=0.75\linewidth]{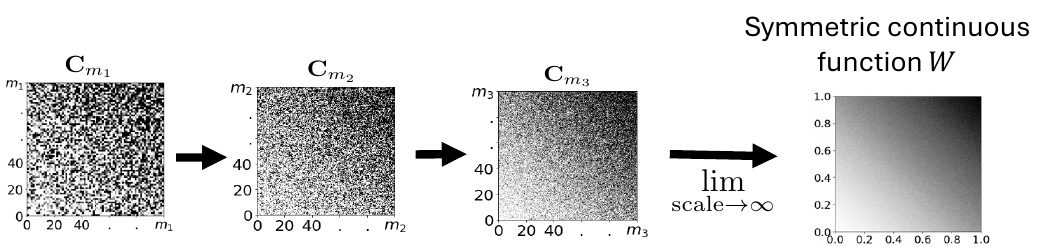}
    \caption{Convergence of a sequence of covariance matrices to a symmetric, continuous function in the limit of scale (number of features) approaching $\infty$.}
    \label{fig:contlimit}
\end{figure}

\begin{figure}
    \centering
    \includegraphics[width=\linewidth]{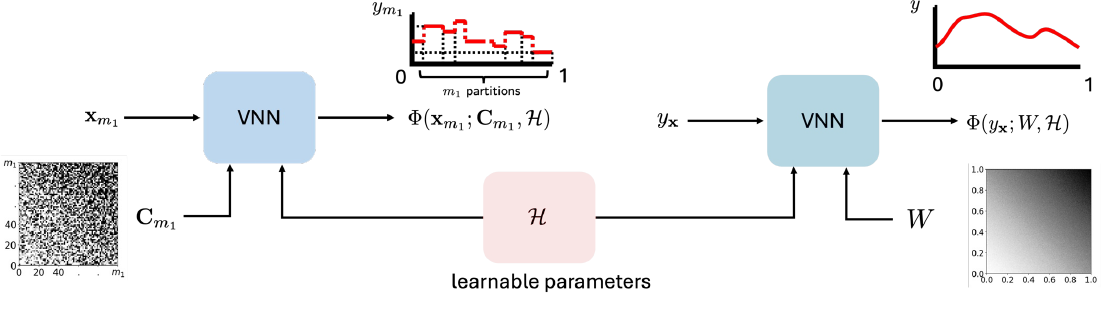}
    \caption{Comparison of representations learned by VNN from a data sample with finite scale and corresponding data sample in the limit of infinite scale.}
    \label{fig:vnntrnsfr}
\end{figure}

\begin{theorem}[Transference of VNNs (Informal)~\cite{sihagJSTSP}]\label{transferthm}
    Consider two datasets of $m_1$ and $m_2$ features and a VNN $\Phi(\cdot;\cdot,\cH)$ consisting of $L$ layers and $F$ outputs per layer. If the continuous approximations $\bW_{\bC_{m_1}}$ and  $\bW_{\bC_{m_2}}$ are close and part of a converging sequence, and the continuous approximations $y_{\bx_{m_1}}$ and $y_{\bx_{m_2}}$ are close, then the continuous approximations of the representations $\Phi(\bx_{m_1};\bC_{m_1},\cH)$ $\Phi(\bx_{m_2};\bC_{m_2},\cH)$ converge as
\begin{align}\label{transfereq}
\|y_{m_1} - y_{m_2}\|_2 = {\cal O} \Big(\frac{1}{m_1^{3\zeta/2- 1} } + \frac{1}{m_2^{3\zeta/2- 1} }\Big)\quad \text{for some constant $\zeta\in (2/3,1]$.}
\end{align}
\end{theorem}
Theorem~\ref{transferthm} implies that continuous representations for the VNN outputs $\Phi(\bx_{m_1};\bC_{m_1},\cH)$ and $\Phi(\bx_{m_2};\bC_{m_2},\cH)$ converge with \blue{an} increase in $m_1$ and $m_2$. As an extension, we expect the measures of central tendency (e.g., mean, median) of the VNN outputs in the multiscale setting to converge as well. Hence, if the VNN provides the inference output with the unweighted mean as the readout function, we expect the statistical outcomes derived from  $\Phi(\bx_{m_1};\bC_{m_1},\cH)$ and $\Phi(\bx_{m_2};\bC_{m_2},\cH)$ to be close or consistent and this convergence to be stronger for large $m_1$ and $m_2$. Key technical components that lead to the results in Theorem~\ref{transferthm} have been discussed in `Analytical components behind the theoretical analysis of transferability of VNNs'. The experiments in Case Study 2 demonstrated the successful transference of VNN models across multiscale neuroimaging datasets curated according to different scales of Schaefer's brain atlas~\cite{schaefer2018local}. 

\begin{mdframed}[hidealllines=true,backgroundcolor=gray!20]
{\bf Analytical components behind the theoretical analysis of transferability of VNNs. }\\ 
The theoretical approach to characterizing transference focuses on defining the \textit{limit of the dataset and covariance network in the asymptote of infinite number of features or dimensionalities}. Assuming that such limits exist, we will be able to characterize the notion of convergence between the datasets at finite scales. The existence of such a limit is intuitively expected for various applications. An example of brain activity being captured on the continuum of the brain surface at different scales in neuroimaging datasets was discussed above. Furthermore, the geospatial datasets capture information about a physical spatial phenomenon, such as ocean temperature or soil properties, via a strategic location of sensors~\cite{cressie2003spatial}.  Figure~\ref{fig:contlimit} provides a figurative illustration of such a sequence, where the covariance matrices formed by datasets with increasing scale or number of features converge to a continuous function ${\bf W}$ as the scale approaches infinity.

The continuous representations of graph signals and graphs have previously been leveraged to study transferability of GNNs under the domain of graphon information processing~\cite{ruiz2021graphon}. Specifically, GNNs can be transferable between graphs belonging to a converging sequence if the graphs in this sequence converge to a limit object called \emph{graphon} as the number of nodes approaches infinity~\cite{borgs2008convergent}. Similarly, the convergence of covariance matrices towards a graphon limit~\cite{borgs2008convergent} and analysis of VNNs within the regime of graphon signal processing~\cite{ruiz2021graphon} is instrumental to establishing Theorem~\ref{transferthm}. Graphons are the limits of \emph{dense} graphs (i.e., graphs with number of edges of the order $\Theta(m^2)$)~\cite{lovasz2012large} and hence, appropriate to study limits of covariance matrices that are typically dense. The definition of a graphon is provided in Definition~\ref{grphon}.
\noindent
\begin{definition}[Graphon]\label{grphon}
A graphon is a bounded, symmetric, measurable function ${\bW: [0,1]^2 \mapsto~[-1,1]}$.   
\end{definition} 
\noindent
Subsequently, the theoretical characterization of transference of VNNs hinges on the following analytical steps:

\noindent
\textbf{Analytical component (a)} \textit{VNNs in the continuous domain}: Defining the VNN in the continuous domain is contingent on demonstrating that a coVariance filter $\bH({\bf C}_m)$ can be equivalently represented using convolution operations over continuous approximations ${\bf W}_{{\bf C}_m}$ of covariance matrix ${\bf C}_m$ and $y_{{\bf x}_m}$ of input sample ${\bf x}_m$. The operation ${\bf C}_m {\bf x}_m$ is fundamental to the convolution operation in $\bH({\bf C}_m){\bf x}$ and therefore, we first demonstrate how its continuous equivalent will be evaluated for the tabular setting. For ${\bf s} = {\bf C}_m{\bf x}_m$, the $i$-th element of ${\bf s}$ is $ [\bs]_i = \sum_{j=0}^m [\bC_m]_{ij} [\bx]_j$. Thus, $[\bs]_i$ is a linear combination of elements in $\bx$ according to the $i$-th row of $\bC_m$. In the continuous space, we can equivalently represent $\bs$ as $ y_{\bs}(u) = \int_{0}^1 \bW_{\bC_m}(u,v) y_{\bx}(v) {\sf d} v$. This observation can be readily extended to define the continuous equivalent of a covariance filter. This is feasible because we can write the entity $\bC_m^k\bx$ in $\bH(\bC)$ in a recursive form. Specifically, if we have $\bs_k = \bC_m^k\bx$, then we can rewrite $\bs_k$ as $\bs_k = \bC_m \bs_{k-1}$, where $\bs_0 = \bx$. Thus, using the same reasoning that established the equivalence between ${\bf s}$ and $y_{\bs}$, we conclude that the continuous representation $y_{\bs_k}$ of $\bs_k$ can be recovered via the following operation $  y_{\bs_k}(u) = \int_0^1 \bW_{\bC_m}(u,v) y_{\bs_{k-1}}(v) {\sf d}v$. Since the coVariance filter output $\bz$ is a weighted aggregation of the terms $\bs_k$, we will be able to write its continuous representation $y_{\bz}$ as $ y_{\bz}(u) = \sum_{k=0}^K h_k y_{\bs_k}(u)$.
Using the mathematical steps leading up to this point, we have demonstrated that the continuous representation of the covariance filter output $\bz$ can be recovered via the convolution operations over the continuous representation $\bW_{\bC_m}$. The extension of this observation to defining the representation $\Phi({\bf x}_m; {\bf C}_m, {\cal H})$ for a VNN to its continuous representation $y_{\Phi({\bf x}_m; {\bf C}_m, {\cal H})}$  is straightforward as it involves building a neural network architecture defined on covariance filters in the continuous domain~\cite{sihagJSTSP}. 

\noindent
\textbf{Analytical component (b)} \textit{Convergence of the sequence of continuous approximations $\{{\bW}_{\bC_m}\}_{m= m_0}^{\infty}$ to limit ${\bf W}$}. The existence of the continuous limit $\bW$ for a sequence of continuous approximations  $\{{\bW}_{\bC_m}\}_{m= m_0}^{\infty}$ for covariance matrices from multiscale datasets (for some arbitrary initial scale $m_0$) is contingent on the appropriate construction of the continuous approximations ${\bW}_{\bC_m}$ and an appropriate measure of \textit{distance} between them. Specifically, when (i) continuous approximations ${\bW}_{\bC_m}$ are constructed on the space $[0,1]^2$, such that, the \textit{area} associated with a feature is proportional to its variance, and (ii) the \textit{cut distance} between the successive continuous approximations relative to some metric follows a Cauchy sequence~\cite{borgs2008convergent}, we can leverage the graphon theory for weighted graphs to conclude that the sequence of the continuous approximations for covariance matrices for the multiscale datasets are indeed converging~\cite{sihagJSTSP}. Intuitively, such a convergence of the continuous approximations across scales implies that the multiscale datasets contain only the redundant information across all scales. 

\noindent
\textbf{Analytical component (c)} \textit{Spectral analysis of VNNs in continuous domain.} The comparison of representations learned by VNNs applied on datasets of different scales will be facilitated by spectral decomposition of the continuous approximations of their respective outputs. Spectral analysis of VNNs in the continuous domain will hinge on the continuous approximations $\bW_{\bC_m}$ and the graphon limit function $\bW$ being \textit{Hilbert-Schmidt operators}~\cite{gohberg1990hilbert}. The implication of this assumption is that the continuous approximations $\bW_{\bC_m}$ and the limit function $\bW$ will be continuous, compact and admit the properties (i) $\int_{0}^1\int_{0}^1 \bW(a,b) {\sf d} a{\sf d} b < \infty$ and $\int_{0}^1\int_{0}^1 \bW_{\bC_m}(a,b) {\sf d} a{\sf d} b < \infty$, and (ii) eigendecompositions of the form:  $\bW(u,v) = \sum\limits_{i\in \mZ\backslash \{0\}} \eta_i \Gamma_i(u)\Gamma_i(v)$, where $\eta_i, \forall i\in \mZ\backslash\{0\}$ are eigenvalues and  $\Gamma_i$ are the eigensignals of $\bW$. Similar properties hold for continuous approximations $\bW_{\bC_m}$ under the assumption of them being Hilbert-Schmidt operators. The consequence of this discussion here is that we can analyze the output of a covariance filter defined on the continuous function ${\bf W}$ as $y(u) = \sum\limits_{i\in \mathbb{Z}\backslash 0}\sum_{k=0}^K h_k \eta_i^k \Gamma_i(u) \int_0^1 \Gamma_i(v) x(v) {\sf d}v$ for an input function $x : [0,1]\mapsto \mathbb{R}$ and compare it mathematically with the spectral decomposition of the covariance filter defined on continuous approximation $\bW_{\bC_m}$~\cite{sihagJSTSP}. 
\end{mdframed}

\begin{figure}
    \centering
    \includegraphics[width=0.75\linewidth]{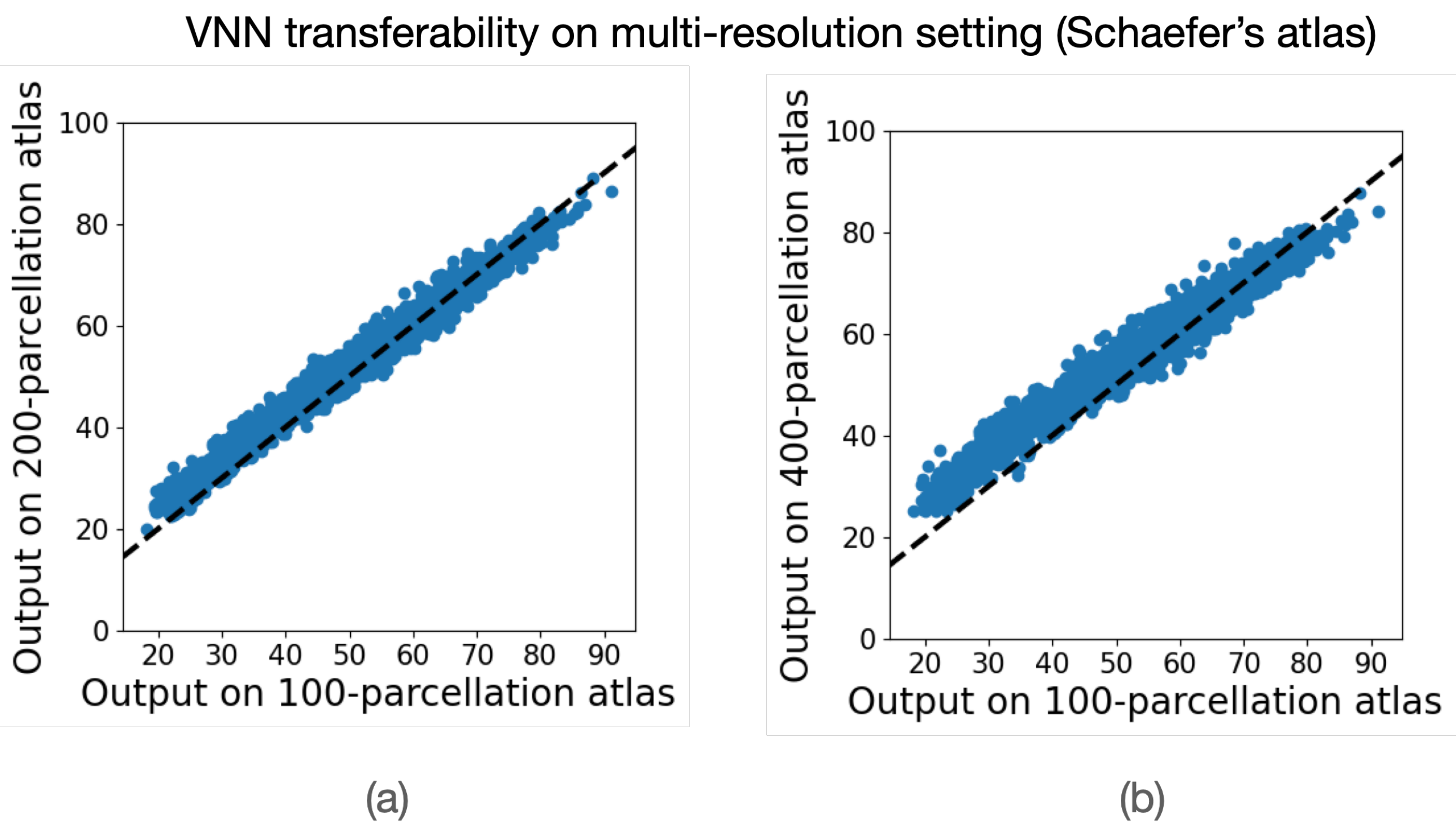}
    \caption{ Transferability of VNN for chronological age prediction task. VNN model was trained on a dataset curated according to the 100-parcellations version of Schaefer's brain atlas and transferred to other versions of the dataset curated according to the 200- and 400-parcellations versions of Schaefer's brain atlas. \blue{Panels (a) and (b) depict the age estimates derived from the VNN on 100-dimensional dataset ($x$-axis) versus the corresponding estimates derived from the same VNN on 200-dimensional version of the same dataset ($y$-axis). The black dotted line represents the line of perfect alignment. }}
    \label{fig:vnntransfr}
\end{figure}

\begin{mdframed}[hidealllines=true,backgroundcolor=gray!20]
\textbf{Case Study 3. Transference of VNNs across multiscale neuroimaging datasets.}

\noindent In this case study, we investigate the transferability of VNN models across multiscale cortical thickness datasets curated according to different scales of 17-network Schaefer's brain atlas~\cite{schaefer2018local}. A VNN model was set up as a scale-free regression model to predict chronological age from cortical thickness features. The scale-free characteristic was achieved with a non-learnable readout function that evaluated the age estimate as the unweighted mean of the outputs at the final layer of the VNN. The VNN was trained on the 100-parcellations dataset to predict the chronological age of a healthy population. Transference of VNN was investigated by deploying the trained VNN model on the dataset from the same population that was curated according to the 200-parcellation and 400-parcellation versions of Schaefer's brain atlas. 

\noindent
{\bf Datasets.} We used the dataset derived from the same healthy population as in Case Study 2. The T1w MRI images in these datasets were preprocessed via the open-source CAT12 pipeline~\cite{gaser2022cat} to yield cortical thickness measures at different scales of Schaefer's atlas for each individual. For every individual, we obtained cortical thickness measures curated according to 100 parcellations, 200 parcellations, and 400 parcellations versions of Schaefer's brain atlas~\cite{schaefer2018local}. The VNN model was trained on the cortical thickness dataset that was curated according to the 100 parcellation version of Schaefer's brain atlas.

\noindent
{\bf Architecture and training.} VNN was trained to predict the chronological age using $100$ cortical thickness features across the cortex for the population of healthy individuals, described in Table~\ref{table_demo}. The architecture consisted of $3$-layers, with $2$ filter taps each in the first and second layers, and $9$ filter taps in the third layer. In contrast to Case Study 2, the readout function was non-learnable. The width of the network was set to $35$. The architecture parameters were selected based on the results of a hyperparameter search performed over $100$ trials using the Optuna package~\cite{akiba2019optuna}. The training procedure was based on optimizing the \blue{MSE} loss function. The learning rate for the Adam optimizer was $0.01$. The model was trained on $90\%$ of the dataset and tested on the remaining $10\%$ to validate that it could infer information about chronological age on an unseen healthy population. The $90/10$ split of the dataset was determined randomly. The model operated on the anatomical covariance matrix of size $100\times 100$ estimated from the $100$ cortical thickness features of the $90\%$ split and normalized, such that its maximum eigenvalue was $1$. Furthermore, the cortical thickness features were z-score normalized across the $90\%$ split and this normalization was applied to the $10\%$ validation cohort. Based on $10$ models trained on different permutations of the training set, VNN achieved an error performance of $9.24\pm0.59$ years with correlation between the ground truth and VNN estimates of chronological age being $0.79\pm 0.004$ on the test set. On the complete dataset, VNN achieved similar performance; the error was $9.16\pm 0.26$ years, and the correlation between the ground truth and VNN estimates of the chronological age was $0.796\pm 0.0032$. 

\noindent
\blue{{\bf Transference of VNN.} The VNN model trained on the $100$-cortical thickness features dataset was subsequently transferred to predict age from $200$-and $400$-features versions. The transference of the VNN model to estimate age from $200$ or $400$ feature datasets was achieved by replacing the covariance matrix with the corresponding sample covariance matrices of size $200\times 200$  or $400\times 400$. Since the readout function was unweighted mean of the final layer outputs of the VNN, the VNN-based information processing pipeline could process data of any arbitrary dimensionality. The success of VNN transference was quantified by the assessment of similarities between the VNN performance for the $100$-dimensional dataset that it was trained on and that on the $200$ or $400$-dimensional datasets.} \blue{We anticipated the VNN transference to be successful as the sample covariance matrices derived from $100$-, $200$-, and $400$-dimensional datasets adhered to the convergence condition for successful transference in Theorem~\ref{transferthm}. Specifically, the the cut distance between the consecutive members
of the series $\{\bC_{100},\bC_{200},\bC_{400}\}$ reduced with an increase in the number of parcels, i.e., $\delta_{\square}(\bC_{100}, \bC_{200}) = 0.005$ and $\delta_{\square}(\bC_{200}, \bC_{400}) = 0.0025$, where $\delta_{\square}$ denotes the cut distance~\cite{borgs2008convergent} and was evaluated using {\sf cutnorm} package in python. These observations were consistent with the properties of a Cauchy sequence in the cut distance metric. The convergence of a series of anatomical covariance matrices corresponding to different scales of Schaefer's brain atlas in terms of cut distance was consistent with the previous results reported in~\cite{sihagJSTSP} for cortical thickness features curated according to different scales of Schaefer's atlas for a different population.   }

\noindent
{\bf Results.} Figure~\ref{fig:vnntransfr} demonstrates high similarity between the VNN-driven age predictions for the 100-dimensional dataset (that it was trained upon) and its outputs after being transferred to the 200 and 400 dimensional versions of the same dataset (Pearson's correlation $>0.97$ for all scenarios).

\noindent
{\bf Remarks.} The results here have demonstrated the transferability of VNNs to datasets of different dimensionalities for the downstream tasks. This ability of VNN is remarkable as it demonstrates the ability of VNN to exploit the redundant information across datasets of different dimensionalities; adding a key perspective to the generalization of VNN. Prior works have exploited the transference of VNNs to demonstrate generalizations of anatomical patterns reflecting neurodegeneration across different neuroimaging datasets~\cite{sihag2023explainable, sihag2025disentangling}. 
\end{mdframed}

\section*{Extensions of VNNs}

The properties of VNNs have established them as a versatile foundation for covariance-based learning. Several recent developments have extended the framework to accommodate complex data structures~\cite{cavallo2024spatiotemporal}, specialized estimators~\cite{cavallo2024fair,cavllo2025robustvnn}, and diverse operational constraints such as label scarcity~\cite{cavallo2025covscatt}, often drawing powerful parallels with classical PCA variants and while investigating the impact of the finite-sample stability.

\subsection*{Task-Based Covariance Estimation}

Building on the foundations of VNNs, several extensions have been developed to address specific data structures or dependencies, often drawing parallels to specialized PCA variants. We describe here the extensions to temporal, biased, and noisy data.

While standard VNNs effectively capture spatial relationships, the SpatioTemporal VNN (STVNN) framework introduced in \cite{cavallo2024spatiotemporal} is designed to account for the intrinsic temporal dependencies found in domains like neuroimaging, sensor networks, and finance. To achieve this, STVNNs modify the standard graph convolution operation by integrating a summation over a temporal window. Consider a dataset of temporal observations $\{ \bx_1, \bx_2, \dots, \bx_n \}$. A spatiotemporal covariance filter with window size $t$ applied to the $n$-th observation performs the operation
\begin{equation}
    \bz = \sum\limits_{\tau=0}^{t-1} \sum\limits_{k=0}^K h_{k\tau} \hat\bC_n^k\bx_{n-\tau}.
\end{equation}
Since the learnable coefficients $h_{k\tau}$ depend both on the time lag and the shift order, this filter effectively captures spatiotemporal patterns.
To support real-world streaming applications, STVNNs utilize online updates for both the covariance estimates and the filter coefficients. Under stationary conditions, this iterative approach guarantees convergence to the true underlying covariance; in non-stationary environments, it enables the model to adapt to distribution shifts. This formulation establishes a direct parallel with spatiotemporal and online PCA variants through the spectral processing of spatiotemporal covariance information. By leveraging the inherent robustness of the VNN architecture, STVNNs provide a mathematically grounded framework that guarantees stability against finite-sample estimation errors.

In~\cite{cavallo2024fair}, VNNs are extended to the domain of algorithmic fairness, ensuring equitable performance across diverse demographic or categorical groups. Specifically, Fair VNNs (FVNNs) aim to achieve consistent accuracy on downstream tasks regardless of group membership, particularly when the task itself should be agnostic to such attributes. Standard sample covariance estimators often inherit and amplify data biases, resulting in PCA projections that favor some groups over others. Moreover, the stability issues intrinsic in PCA yield principal components that are more accurate for majority groups and more noisy for underrepresented ones, amplifying the existing biases. 
The FVNN paradigm addresses this problem with two improvements. First, FVNNs employ fair covariance estimators, that either reweigh the contributions of different groups to compensate for sample size imbalances or apply transformations to decouple group dependencies from the covariance structure entirely. Second, FVNNs are trained to minimize an objective function that balances the downstream task performance and a bias mitigation penalty. This formulation effectively extends fair PCA methods by introducing learnable, fair weights, providing greater flexibility in capturing task-relevant information. Furthermore, FVNNs inherit the characteristic stability of the VNN architecture, remaining robust to finite-sample estimation errors within the fair covariance estimators themselves and mitigating unfair treatments related to covariance estimation errors.

Finally, the work in~\cite{cavllo2025robustvnn} considers the setting where observations are perturbed by outliers (e.g., interferences in sensor measurements) or missing values (e.g., temporary sensor failures). Such perturbations differ from the finite-sample estimation errors as they are significantly larger in magnitude, thus violating the assumptions behind VNNs' stability result and deteriorating their downstream task performance. 
To mitigate these effects, Robust VNNs (RVNNs)~\cite{cavllo2025robustvnn} estimate two covariance correction matrices that compensate for data corruption. Following the robust PCA data model~\cite{candes2011robust}, these corrections are modeled using one low-rank and one sparse prior, and they are learned end-to-end by minimizing a loss that balances downstream task accuracy with the structural priors of the correction matrices. In this way, RVNNs produce task-aware robust covariances that filter out large-scale perturbations while simultaneously leveraging label information. This integrated approach ensures that RVNNs remain stable in the presence of both standard finite-sample estimation errors and significant data corruption.

\subsection*{Covariance Scattering Transforms}

VNNs are inherently supervised architectures that rely on training to estimate an appropriate set of coefficients, which generally requires labeled data for a downstream task. 
In many practical applications, however, data is abundant but labels are scarce, rendering standard VNNs difficult to deploy.
To overcome this limitation, Covariance Scattering Transforms (CSTs)~\cite{cavallo2025covscatt} provide a framework for generating robust, covariance-aware representations without training.
CSTs are based on covariance wavelets, filtering functions localized in both the spectral and spatial domains of the covariance matrix. These wavelets extract specific covariance patterns and are instantiated at different scales to capture shifted or modified versions of these patterns. 
CSTs consist of a cascade of covariance wavelets at different scales applied to the input and interleaved with non-linear activation functions. This multi-layer architecture produces data representations that are covariance-aware, expressive, untrained, and stable to finite-sample estimation errors, thus merging the untrained nature of PCA and the stability and expressivity of VNNs. 
Since CSTs' representations can significantly grow in dimension, a pruning mechanism is adopted to remove low-energy branches, speed up computation, and reduce memory consumption while maintaining stability and expressivity.
Ultimately, the high-quality features produced by CSTs allow simple readout models to achieve strong downstream task performance even in low-data regimes or settings with minimal label availability.

\section*{Conclusions and Future Outlook}
PCA has been widely adopted for data analysis and statistical inference in recent decades. The success of PCA stems from various factors, including minimal assumptions, model-agnostic characteristics, and interpretability tied to intuitive understanding of the covariance matrix. We began this tutorial by highlighting the critical conceptual and operational challenges faced by the conventional covariance matrix-based PCA approaches; broadly encompassing the lack of scalability (due to their dependence on explicit eigendecomposition of the covariance matrix) and lack of guarantees on the reproducibility of findings in finite data and multiscale data regimes. The aforementioned challenges appear prominently in modern data regimes characterized by high-dimensional data samples and complex data acquisition protocols. While the aforementioned shortcomings to PCA have been recognized individually in prior works, there has been a lack of holistic update to the principles of PCA that prominently addresses them in an integrated fashion. In this tutorial, we have aimed to address this gap through a GSP perspective to PCA, which stems from the graphical interpretation of a covariance matrix. Specifically, a covariance matrix admits a graphical interpretation with features of the given dataset as the nodes of the graph and the pairwise covariances informing the edge structure. 

Our starting point was the direct theoretical equivalence between a PCA-based learning algorithm that assigns weights to individual principal components and a graph filter implemented on a covariance matrix as the graph. This equivalence demonstrated an alternative implementation of a PCA-based learning model in terms of a polynomial over the covariance matrix, which offers several key benefits over the conventional PCA-based learning models: (a) graph filter implementation of PCA does not require an explicit eigendecomposition of the covariance matrix; (b) graph filters form a task-specific and parametric learning model unlike PCA, which is commonly adopted as a pre-processing step and decoupled from the learning task; and (c) scale-free characteristic implementation of graph filters allow them to be amenable to processing multiscale information with the same model, unlike PCA which is tied to the dimensionality of the given dataset. The expressive power of a graph filter could be enhanced by adding pointwise non-linearities and building banks of graph filters, leading towards VNN being defined as the GNN with a covariance matrix as the graph.

The surveyed theoretical results in this article rigorously demonstrated the robustness of VNN outcomes to stochastic perturbations in the covariance matrix as well as transference across datasets of different scales. Robustness of learning outcomes to stochastic perturbations in the covariance matrix enforces their reproducibility in finite-data regimes. This is because VNNs operate over the sample covariance matrix in practice, and the said robustness guarantees consistent learning outcomes across different sample covariance matrices from other finite-sized datasets sampled from the same distribution. Theoretical guarantees on the transference across multiscale datasets imply the effectiveness of VNNs at leveraging the redundancy of signals at different resolutions, which can potentially lead to computation- or data-efficient ML frameworks. While these theoretical guarantees reinforced the categorical advantages brought in by the GSP perspective to PCA, our aim here was not to diminish the PCA transform or put VNNs in contrast to it, but rather to offer significant conceptual advancements to PCA through VNN models. Achieving this goal involved bridging PCA with graph filters by demonstrating natural analytical and conceptual connections.

GSP-driven learning architectures have previously been shown to be robust to abstract perturbations in the graph as well as transferable across graphs of different sizes. While these properties extend naturally to VNNs, this tutorial article has also highlighted the unique insights brought in by the data-driven construction of the covariance matrix and the perturbation theory governing the divergence between the sample covariance matrix and its true counterpart. Specifically, we discussed the refined theoretical bounds in the context of VNNs, which enabled the understanding of their dependence on the data size and specific properties of the covariance matrix. GNNs are widely implemented on covariance matrices, and the theoretical results surveyed here contribute to their principled deployments in regimes with finite data. Furthermore, the discussions on various extensions of the VNNs to settings with fairness considerations, temporally evolving datasets, and corrupted data highlight the broad relevance of the theoretical principles surveyed herein to applications where covariance matrices emerge and PCA has been a workhorse analytical tool.  

Looking ahead, this article has set the groundwork for enhancing the theoretical concepts of other multivariate statistical inference methods besides PCA as well as modern deep learning models that leverage covariances. For instance, a recent study of graph filters from the lens of neural tangent kernels~\cite{jacot2018neural} has revealed cross-covariance between the inputs and the desired outputs to contribute to the optimal generalization and training performance of a GNN~\cite{khalafi2024neural}. This observation sets up a potential fundamental connection between a GNN with cross-covariance graphs and multivariate statistical methods, such as canonical correlation analysis (CCA), that explicitly leverage cross-covariance. In this context, future studies that address the shortcomings of conventional statistical methods (for instance, related to instability of CCA~\cite{helmer2024stability}) are well-motivated. Moreover, covariances are also a significant part of various prevalent deep learning architectures, such as transformers. The conceptual similarities between the self-attention mechanism and coVariance filters (highlighted in `Self-Attention in Transformers versus CoVariance Filter in VNNs') reveal a promising direction towards theoretical understanding of transformers. 

Furthermore, the foundational advances surveyed in this article can promisingly offer benefits in science and engineering applications where covariance matrices emerge. A recent tutorial article highlighted the key points of clarity and advancements offered by VNNs to the brain age gap prediction in the computational neuroscience domain~\cite{sihag2025disentangling}. Specifically, the VNN-driven brain age gap prediction pipeline offered a promising alternative to the currently prevalent ML approaches in this application that are largely opaque and performance-driven; and hence, not practical for clinical deployment. VNNs offer key benefits in this context due to enhanced interpretability and guaranteed stability and generalizability of anatomical patterns characterizing neurodegeneration. Furthermore, multiscale information is increasingly prevalent across various applications, For instance, the physical processes in atmospheric or oceanographic sciences are typically monitored with surface-level, airborne, and satellite-level sensors~\cite{neelam2020multiscale}; thus, revealing the need for learning models that can process multiscale information. Multiscale information is appealing in ML because it encodes the redundancy of information at different scales (spatial or temporal), potentially leading to computation- and sample-efficient training mechanisms and deployment via transference from small- to large-scale datasets. While recent works have proposed various deep learning algorithms that can succeed in inference tasks across multiscale datasets in specific domains, the mathematical principles behind such models are lacking and could be enhanced with the foundational advancements behind VNNs.

In summary, VNNs hold great promise to drive the necessary conceptual updates to conventional statistical methods to meet the needs of modern data analysis and enhance the theoretical understanding of modern deep learning methods. We argue that VNNs are particularly useful in finite-data regimes, where their theoretical guarantees ensure consistent and reliable learning outcomes. Furthermore, the theory of VNNs can lend unparalleled depth while offering major conceptual advancements in the design of application-relevant data analysis solutions.

---------------------------------------------------
%
\section*{Acknowledgment}
CamCAN dataset was obtained from the CamCAN repository~\cite{taylor2017cambridge,shafto2014cambridge}. Dallas Lifespan Brain Study (DLBS) is available at International Neuroimaging Data-sharing Initiative (INDI). IXI dataset is available at \href{https://brain-development.org/ixi-dataset/}{https://brain-development.org/ixi-dataset/}. The eNKI dataset is available at\\ \href{https://fcon\textunderscore 1000.projects.nitrc.org/indi/pro/eNKI\textunderscore RS\textunderscore TRT/FrontPage.html}{https://fcon\textunderscore 1000.projects.nitrc.org/indi/pro/eNKI\textunderscore RS\textunderscore TRT/FrontPage.html}.

---------------------------------------------------
%

\bibliographystyle{IEEEtran}
\bibliography{spm_bib}

%
%
%
%
%
\section*{Biographies}
\label{sec:bio}
%
%

\begin{IEEEbiographynophoto}{Saurabh Sihag} is an Assistant Professor in the Department of Electrical and Computer Engineering at the University at Albany. He received his PhD degree in Electrical Engineering from Rensselaer Polytechnic Institute, in 2020, and his Bachelor’s and Master’s degrees in Electrical Engineering from Indian Institute of Technology, Kharagpur in 2016. He was a postdoctoral researcher in the Department of Electrical and Systems Engineering at the University of Pennsylvania between March, 2021 and July, 2024. He has previously been the recipient of J. Baliga fellowship and Charles M. Close ’62 Doctoral Prize for his doctoral dissertation. His research interests include statistical inference and machine learning over graph-structured data and network neuroscience. 
\end{IEEEbiographynophoto}
%

\begin{IEEEbiographynophoto}{Andrea Cavallo} received his B.Sc. and M.Sc. degrees (cum laude) in Computer Engineering from the Polytechnic University of Turin, Italy, in 2020 and 2022, respectively. Since 2023, he is a Ph.D. candidate at the Faculty of Electrical Engineering, Mathematics and Computer Science, Delft University of Technology, Netherlands. His research focuses on relational machine learning through higher-order networks and statistical moments as well as temporal network modeling and prediction.
\end{IEEEbiographynophoto}

\begin{IEEEbiographynophoto}{Elvin Isufi} is an Associate Professor in the Faculty of Electrical Engineering, Mathematics and Computer Science at Delft University of Technology (TU Delft), The Netherlands, where he also co-directs AIdroLab –the TU Delft AI Lab on graph-based learning for water networks. He received the Ph.D. degree from the same university (’19), and the M. Sc. (’14) and B. Sc. (’12) in Electronic and Telecommunication Engineering from the University of Perugia, Italy. His research interests are in signal processing and machine learning for graphs and other topological structures with applications in critical infrastructure networks, recommender systems, and sensor networks. He received the 2022 IEEE SPS Best PhD Dissertation Award, the 2025 ICASSP Best Conference Paper Award and paper recognition awards at the IEEE CAMSAP (’17), DSLW (’21, ’22), and ICASSP (’23). He is a member of the IEEE SPS Technical Committee on Signal Processing for Communication and Networking and serves as Associate Editor for the IEEE Transactions on Signal Processing, position that he also had held for Elsevier Signal Processing (’24-’26). He is an NWO VENI fellow (’24) and a Horizon MSCA Cofound fellow (’19).
\end{IEEEbiographynophoto}

\begin{IEEEbiographynophoto}{Gonzalo Mateos} received his B.Sc. degree in Electrical Engineering from Universidad de la Rep\'ublica, Montevideo, Uruguay in 2005 and the M.Sc. and Ph.D. degrees in Electrical Engineering from the University of Minnesota, Minneapolis, in 2009 and 2012. Currently, he is a Professor with the Department of Electrical and Computer Engineering, University of Rochester, as well as the Associate Director for Research at the Goergen Institute for Data Science and Artificial Intelligence. He also was an Asaro Biggar Family Fellow in Data Science (2020-23). His research interests lie in the areas of statistical learning from complex data, network science, decentralized optimization, and graph signal processing.
\end{IEEEbiographynophoto}

\begin{IEEEbiographynophoto}{Alejandro Ribeiro} received the B.Sc. degree in Electrical Engineering from the Universidad de la Rep\'ublica, Montevideo, Uruguay, in 1998, the M.Sc. and Ph.D. degrees in Electrical Engineering from the University of Minnesota, Minneapolis, MN, USA, in 2005 and 2007, respectively. Since 2008, he has been with the University of Pennsylvania (Penn), Philadelphia, PA, USA, where he is currently a Professor of Electrical and Systems Engineering. His research interests include applications of statistical signal processing to the study of networks and networked phenomena, structured representations of networked data structures, graph signal processing, network optimization, robot teams, and networked control. Dr. Ribeiro was the recipient of an Outstanding Research Award from Intel in 2019, the 2014 O. Hugo Schuck Best Paper Award, and paper awards at ICASSP 2020, EUSIPCO 2019, CDC 2017, 2016 SSP Workshop, 2016 SAM Workshop, 2015 Asilomar SSC, ACC 2013, ICASSP 2006, and ICASSP 2005. His teaching has been recognized with the 2017 Lindback Award and the 2012 S. Reid Warren, Jr. Award presented by Penn’s undergraduate student body for outstanding teaching. Dr. Ribeiro is a Fulbright Scholar class of 2003 and a Penn Fellow class of 2015.
\end{IEEEbiographynophoto}

\vfill

\end{document}